\documentclass[acmsmall,screen,review=false,table]{acmart}

\AtBeginDocument{%
  \providecommand\BibTeX{{%
    \normalfont B\kern-0.5em{\scshape i\kern-0.25em b}\kern-0.8em\TeX}}}
\usepackage{booktabs} % For formal tables
\usepackage{epsfig}
\usepackage{amsmath}
\usepackage{amsfonts}
\usepackage{xcolor}
\usepackage{algorithm}
\usepackage[noend]{algpseudocode}
\usepackage{multirow}
\usepackage{mathtools}
\usepackage{subcaption}
\usepackage{wrapfig}
\usepackage{tabulary}
\usepackage{colortbl}
\usepackage{tikz}
\usepackage{tcolorbox}

\newcommand{\greyball}[1]{%
  \tikz[baseline=(char.base)]{
    \node[shape=rectangle, rounded corners, fill=gray, text=white, inner sep=3pt] (char) {#1};
  }%
}

\usepackage{enumitem}

\newcommand{\subsubfour}[1]{\vspace*{1mm}{\noindent\bf #1}}

\usepackage{pifont}

\newcommand{\cmark}{\textcolor{green}{\ding{51}}} % Green check mark
\newcommand{\xmark}{\textcolor{red}{\ding{55}}}   % Red cross

\author{Brandon Smith}
\affiliation{
  \institution{Deakin University}
  \city{Burwood}
  \state{VIC}
  \country{Australia}
  }
\email{brandon.smith@deakin.edu.au}

\author{Mohamed Reda Bouadjenek}
\affiliation{
  \institution{Deakin University}
  \city{Geelong}
  \state{VIC}
  \country{Australia}
  }
\email{reda.bouadjenek@deakin.edu.au}

\author{Tahsin Alamgir Kheya}
\affiliation{
  \institution{Deakin University}
  \city{Geelong}
  \state{VIC}
  \country{Australia}
  }
\email{t.kheya@deakin.edu.au}

\author{Phillip Dawson}
\affiliation{
  \institution{Deakin University}
  \city{Melbourne}
  \state{VIC}
  \country{Australia}
  }
\email{p.dawson@deakin.edu.au}

\author{Sunil Aryal}
\affiliation{
  \institution{Deakin University}
  \city{Geelong}
  \state{VIC}
  \country{Australia}
  }
\email{sunil.aryal@unsw.edu.au}

\AtBeginDocument{%
  \providecommand\BibTeX{{%
    Bib\TeX}}}

\setcopyright{acmcopyright}
\acmJournal{TWEB}
\acmYear{2025} 
\acmVolume{1} 
\acmNumber{1} 
\acmArticle{1} 
\acmMonth{1} 
\acmPrice{15.00}
\acmDOI{10.1145/3532856}    

\begin{document}
% \fancyhead{}
% \fancyhead[RE]{}

\title{A Comprehensive Analysis of Large Language Model Outputs:  Similarity, Diversity, and Bias}
% Evaluating Large Language Model Outputs: A Study on Similarity, Diversity, and Bias
% \title{A Comparative Analysis of Large Language Models: Exploring Writing Similarities and Variances }

\begin{abstract}
% Context of work
Large Language Models (LLMs) represent a significant leap forward in the quest for artificial general intelligence, enhancing our capacity to engage with and leverage technology. 
However, while LLMs have demonstrated their effectiveness in various Natural Language Processing tasks such as language translation, text generation, code generation, and content summarization, among others, there remain many open questions about their similarities, their variance, and their ethical implications.
% Research questions to address in the paper
For example, when presented with a text generation prompt, what similarities exist among texts produced by the same language models? 
In addition, what is the inter-LLM writing similarity - for instance, how comparable are texts generated by distinct LLMs when presented with the same prompt?
Furthermore, how does the variation in text generation manifest across multiple LLMs, and which LLMs adhere most closely to ethical standards?
% Data used
To address these questions, we used 5K different prompts covering a diverse range of requests, including generating, explaining, and rewriting text. 
This effort resulted in the generation of approximately 3M texts from 12 different LLMs, featuring a mix of proprietary and open-source models from industry leaders such as OpenAI, Google, Microsoft, Meta, and Mistral.
% Insights on results
% The study reveals significant insights, including the observation that text generated by identical language models typically demonstrates greater consistency and similarity compared to human-written text.
The results of this study reveals a number of important insights, among them that:
(1) texts produced by the same LLMs show higher similarity compared to human-written texts;
(2) some LLMs, like WizardLM-2-8x22b, produce highly similar texts, while others like GPT-4 generate more varied outputs;
(3) writing styles among LLMs vary significantly, with models like Llama 3 and Mistral showing high similarity, while GPT-4 stands out for its distinctiveness;
(4) the sharp contrast in language and lack of vocabulary overlap highlight the distinct linguistic characteristics of LLM-produced text;
(5) finally, it appears that certain LLMs are more balanced in terms of gender representation and are less prone to perpetuating bias.


\end{abstract}

% \keywords{Large Language Models;
% Natural Language Processing;
% Text Generation}

\maketitle

% \vspace{-0.25cm}
% \noindent
% {\bf Keywords:} Hierarchical Classification, CNN, Deep Learning.

\section{Introduction}
% This is an example of a grey ball with the number \greyball{RQ1} in it. You can create more balls like \greyball{2}, \greyball{3}, and so on.

% Context of work
In the past year, there has been a significant stride in the pursuit of artificial general intelligence, mainly highlighted by the implementation of Large Language Models (LLMs)~\cite{gunasekar2023textbooks,jiang2024mixtral,geminiteam2023gemini,jiang2023mistral,almazrouei2023falcon,touvron2023llama,anil2023palm,openai2024gpt4} to crafting chatbots such at ChatGPT, BARD, Bing Chat, and Grok. These models showcase unparalleled precision and accuracy, particularly in excelling at diverse Natural Language Processing (NLP) tasks, including but not limited to, language translation, text generation, code synthesis, content summarization, conversation, and information search.

% Problem
While these LLMs and chatbots undoubtedly help users complete their tasks, many open questions remain concerning ethical considerations, utilization challenges, output quality concerns, and the similarities and variances among them. 
Hence, in this paper, we seek to analyze their performance across diverse prompts, compare their outputs to human-written texts, and investigate the distinctive characteristics and variations in the texts they generate. 
Our fundamental research questions are focused on understanding the intrinsic characteristics of texts generated by LLMs, as outlined below:

% This is an example of a grey ball with the number \greyball{RQ1} in it. You can create more balls like \greyball{2}, \greyball{3}, and so on.

\begin{enumerate}[leftmargin=1em, itemsep=0.6em,label=]
    \item \greyball{RQ1} When presented with a prompt, what similarities exist among texts produced by the same LLM (inner-LLM similarity), and how similar are the texts generated by different LLMs for the same prompt (inter-LLM similarity)?
    \item \greyball{RQ2} How does the variation in text generation manifest across multiple LLMs? 
    \item \greyball{RQ3} Can we accurately identify whether a given text was authored by a human or a specific language model?
    \item \greyball{RQ4} Are there specific words that can act as distinctive markers for each LLM?
    \item \greyball{RQ5}  Are there LLMs that adhere most closely to ethical standards  by reducing the propagation of biased stereotypes?
\end{enumerate}

% For example, 
% In addition, 
% Furthermore, 
% Finally,
% % What we propose in the paper
% To address our research questions, we utilised twelve distinct LLMs, both proprietary and non-proprietary LLMs. Initially, we selected fifteen unique prompts to generate approximately 13,000 texts, supplemented by human-generated responses to the same prompts to allow for comparability and depth of analysis.

% Expanding our dataset further, we extracted 5,000 instructions characterized by keywords such as generate, rewrite, write, and explain from the Alpaca dataset. These instructions were then used to generate texts using the aforementioned models, with each model producing outputs for every instruction 20 times under non-deterministic settings. This approach resulted in a comprehensive dataset of 940,000 texts. However, to better understand the capabilities of language models in specific contexts, particularly in literary and creative writing we pruned our initial dataset by concentrating our efforts on instructions that explicitly involved tasks such as generating stories, rewriting texts, and other writing-focused instructions highlighted in Figure~\ref{fig:dataset_keywords}. By applying these criteria, we refined the dataset down to 360,000 texts. The insights from this analysis highlight several key findings:

Addressing these questions is crucial as it will provide us with deeper insights into the operational challenges and performance nuances of LLMs and Chatbots across diverse domains, including education, scientific writing, and business communication.
Hence, we propose in this paper a comprehensive analysis of LLM outputs. 
In particular, we employed approximately 5,000 distinct and diverse prompts covering diverse topics ranging from technological impact to academic performance.
These prompts encompass a wide range of requests, including text generation, explanation, and rewriting.
Using these prompts, we generated texts with 12 different LLMs, including proprietary and open-source models such as: 
Gemini-pro-1.5~\cite{geminiteam2023gemini}, 
Gemma-7B~\cite{gemmateam2024gemma},
GPT-3.5, 
GPT-4~\cite{openai2024gpt4}, 
Mistral-7B, 
Mixtral-8x7B, 
Mixtral-8x22B~\cite{jiang2023mistral},
WizardLM-2-7B,
WizardLM-2-8x22B~\cite{xu2023wizardlm},
Llama 3-70B,
Llama 3-8B~\cite{touvron2023llama},
and DBRX.
Additionally, we include human-generated text produced using 15 prompts as instructions for text comparison.
The resulting dataset comprises 3 Million texts and is utilized to conduct a thorough analysis from which we draw meaningful conclusions including:
% \vspace{-\topsep} % Adjust the space before the list
\begin{itemize}[leftmargin=1em, itemsep=0.6em]

\item \textbf{Low Similarity within LLMs:} The texts produced by the same LLMs generally show higher consistency and similarity in their outputs compared to human-written texts. Some LLMs, such as WizardLM-2-8x22b, produce highly similar texts, while others like GPT-4 generate more varied outputs.
Additionally, proprietary models tend to be more consistent than open-source models in terms of output similarity.
\item \textbf{Inter-LLM Writing Similarity:} The writing styles vary significantly, with some models like Llama 3 and Mistral showing high similarity, while GPT-4 stands out for its distinctiveness. 
Although at the same time word-level similarity measures indicated that GPT-4 was the most distinct and not similar to GPT-3.5, BERT revealed a different aspect, showing that GPT-4 and GPT-3.5 are similar in deeper, contextual ways. 
\item \textbf{Variance in Text Generation:} Some LLMs demonstrated significant variance in text generation, while others exhibited a more consistent output. 
This insight into the diversity of LLM behavior is crucial for understanding the nuances of these models.
\item \textbf{Classification:} Our classification efforts show success in being able to differentiate between human-written text and text generated by various LLMs. 
% BERT achieved the highest multi-label accuracy and outperformed all other models, including DeBERTa variants and XGBoost-BoW. 
Misclassifications and confusion mainly occurred between similar models like GPT-3.5 and GPT-4, highlighting the challenge of distinguishing between texts from closely related architectures.
\item \textbf{Language markers:} The sharp contrast in language and the absence of vocabulary overlap emphasize the distinct linguistic characteristics between human-generated text and that produced by the language models.
\item \textbf{Ethics consideration:} it seems that certain LLMs like Gemma-7B and Gemini-pro demonstrate a more balanced approach to gender representation and are less likely to perpetuate bias.
In contrast, models such as GPT-3.5 and GPT-4, while powerful in terms of performance, demonstrate a stronger tendency toward gender and racial bias.

% \item Explainability Challenges: Establishing strong evidence for attributing a text to a particular LLM or a human remains a significant challenge. The need for explainability is important, and it remains an open research question on how to achieve it convincingly.
\end{itemize}
% In summary, this paper builds on the complexities and potential of LLMs as highlighted in our comprehensive analysis. 
% We have not only explored the variance in LLM behaviors when responding to identical prompts but also examined the similarities and differences in outputs across a variety of proprietary and non-proprietary models.\

In summary, this paper sheds light on the intricacies of LLM behavior and strives to advance the discourse on the responsible development and utilization of LLMs. 
% Furthermore, this paper contributes a dataset that can be used for further research into LLM behavior, including investigations into the consistency of model outputs, variations in response to different types of prompts, and the ethical implications of automated text generation.

% \vspace{-\topsep} % Adjust the space before the list

\section{Related work}

LLMs have emerged as a prominent subject of research among the academic community. 
Below, we review the evolution of Language Models, including research related to the analysis of LLMs, efforts focused on their detection and classification, and studies addressing bias and discrimination in LLMs.

\subsubfour{Evolution of Language Models:}
% LLMs such as those detailed in \cite{brown2020language, geminiteam2023gemini, openai2024gpt4, almazrouei2023falcon} have been influential in the field of natural language processing (NLP), trained on extensive corpora and utilising hundreds of billions of parameters. 
% These models have been central due to their large scale and range of capabilities, including understanding, creating, and translating natural language \cite{ai4science2023impact}.
The field of NLP has grown rapidly since the introduction of word embeddings in 2013~\cite{NIPS2013_9aa42b31}, with Word2Vec providing a foundation for capturing semantic word relationships in the vector space.
This foundational approach gained further significance when employed in conjunction with sequence models like RNNs and LSTMs  \cite{hochreiter1997long}, establishing itself as a critical element in addressing complex NLP tasks.
% Indeed, their usage with sequence models such RNNs and LSTM architecture \cite{hochreiter1997long} became a centrepiece for complex NLP tasks.
% The introduction of RNNs and the LSTM architecture \cite{hochreiter1997long} provided further developments in sequential data processing, which addressed the vanishing gradient problem and became a centrepiece for complex NLP tasks.
Additionally, a significant progression was the introduction of the Transformer architecture by Vaswani et al. in 2017~\cite{NIPS2017_3f5ee243}, which moved away from recurrent layers in favor of self-attention mechanisms, leading to parallel processing and a reduction in training times.
Moreover, Google's BERT Language Model, introduced in 2018~\cite{devlin-etal-2019-bert}, employed \textit{encoder} transformer blocks to learn bi-directional contextual representations and set new performance benchmarks across a variety of NLP tasks. 
In contrast, OpenAI introduced the GPT series~\cite{radford2018improving, NEURIPS2020_1457c0d6}, in particular GPT-3 with its 175 billion parameters, based on \textit{decoder} transformer blocks achieved state-of-the-art language generation capabilities \cite{NEURIPS2020_1457c0d6}.
This evolution has launched the quest toward training LLMs that consistently demonstrate a form of near artificial general intelligence across various NLP tasks.

\subsubfour{Analyzing LLMs:}
% LLMs have also introduced significant challenges, particularly in 
% distinguishing between human-written text and text generated by LLMs with the key issue lying in the interpretability and explainability of these models. 
% They are also significantly impacting higher education by offering tools for writing, coding and information retrieval, while raising the ethical concerns of academic integrity~\cite{zhou2024the}. 
While LLMs have proven to be highly useful, their adoption has also introduced significant challenges, particularly in terms of explainability to allow effective debugging and performance enhancement~~\cite{huang2023large, zhao2023survey, zhou2023mystery, 10.1145/3639372}.
% Works such as those by have aimed to explain these complex models in a variety of ways such as their abilities on different tasks and how explainability can be used to debug and improve performance. 
% Hence, recent works have looked at improving the overall capabilities of LLMs from diverse perspectives such as XYZ~\cite{liu2023improving}, XYZ~\cite{nguyen2023enhancing}, and XYZ~\cite{sharma2023truth}.
On the other hand, there are other challenges associated with their usage that need to be analyzed, including:
(1) \textbf{Trustworthiness and Toxicity:} Recent studies~\cite{sun2024trustllm, mo2023trustworthy} have highlighted the issues of trustworthiness and safety in LLM outputs. 
These concerns are worsen by the potential for LLMs to generate toxic content~\cite{wen-etal-2023-unveiling, deshpande-etal-2023-toxicity}. 
Hence, recent work~\cite{prabhumoye-etal-2023-adding} has explored mitigating such risks  by including pre-training instructions.
(2) \textbf{Memorisation:} LLMs tend to memorise and regurgitate training data~\cite{de_Wynter_2023, neel2024privacy}, where privacy can be for instance violated during inference~\cite{staab2023memorization}.
This represents significant challenges related to privacy, utility, and fairness. 
% There has been investigation on how privacy can be violated through inference~\cite{staab2023memorization}.
(3) \textbf{Hallucination and Reliability:} The issue of hallucination in LLM responses~\cite{du2023quantifying, xu2024hallucination} further complicates their reliability. 
A few works have looked at mitigating these hallucinations~\cite{bruno2023insights} by exploring various strategies such as fine-tuning~\cite{Lee2020}, memory augmentation~\cite{wu-etal-2022-efficient}, or prompt
strategies~\cite{10207581}.

% \textbf{Memorisation:} LLMs tend to memorise and regurgitate training data as highlighted in~\cite{de_Wynter_2023, neel2024privacy}. This presents significant challenges related to privacy, utility, and fairness. There has been investigation on how privacy can be violated through inference~\cite{staab2023memorization}.

\subsubfour{Detecting LLM generated text:}
% Our work extends on identifying and detecting outputs from LLMs, particularly within human written text. 
% Related work includes those such as~\cite{peng2024hidding, feng2024does, mao2024raidar, gao2024llmasacoauthor, caiado2023ai, sadasivan2023aigenerated, koike2023prompt, liang2023gpt}.
Detecting text generated by LLMs is a critical task, and several studies have explored this issue. 
These investigations span from assessing the effectiveness of current detection systems against adversarial attacks~\cite{peng-etal-2023-hidding} to the complex challenges of social media bot detection~\cite{feng2024does} and the identification of texts that blend human-written and LLM-generated content~\cite{gao2024llmasacoauthor}. 
However, the reliability of these systems varies significantly when LLMs act as co-authors, influenced by factors such as the user's prompt or the linguistic background of the user.
% or when the text originates from non-English writers. 
% The recent works reveal that the performance of detection systems can vary greatly, a variability that is influenced not only by the structure of user prompts but also by the linguistic background of the text. Such observations highlight the challenges involved in distinguishing between authentic and artificially manipulated texts, highlighting the complexity of accurately identifying LLM-generated content. 
In response to these challenges, several detection methods have been proposed 
% in recent literature~\cite{corizzo2023oneclass, wang2023seqxgpt, abburi2023generative, mireshghallah2023smaller, kumarage2023neural}, 
including one-class models that treat the target text as outliers~\cite{corizzo2023oneclass}, techniques leveraging sentence-level features for classification~\cite{wang-etal-2023-seqxgpt}, ensemble approaches~\cite{abburi2023generative}, and methods utilising stylometric features and smaller models~\cite{mireshghallah2023smaller, kumarage2023neural}.

% \vspace{-0.3cm}
\begin{table}[t]
    \centering
    \caption{Summary of LLMs and their Key Features, with ``\cmark'' indicating the presence of the feature, ``\xmark'' indicating its absence, and ``?'' indicating that the information is not provided.}
    \label{tab:LLM-summary}
    % \vspace{-0.3cm}
    \resizebox{0.65\textwidth}{!}{%
    \begin{tabular}{|c||c|c|c|c|c|c|}
        \hline
        \textbf{} & \textbf{Model} & \textbf{Company} & \textbf{\#Parameters} & \textbf{Open-Source} & \textbf{Training } & \textbf{MoE$^{\dag}$} \\ 
        \textbf{} & \textbf{} & \textbf{} & \textbf{} & \textbf{} & \textbf{ Data Size} & \textbf{ Architecture} \\ \hline\hline
        1 & Gemini-pro-1.5 & Google & ? & \xmark & 10M tokens & \cmark \\ \hline
        2 & Gemma-7B & Google & 7B & \cmark & ? & \xmark \\ \hline
        3 & GPT-3.5 & OpenAI & 175B & \xmark & 570 GB text and code & ? \\ \hline
        4 & GPT-4 & OpenAI & ? & \xmark & 1.2T tokens & ? \\ \hline
        5 & Mistral-7B & Mistral & 7B & \cmark & ? & \xmark \\ \hline
        6 & Mixtral-8x7B & Mistral & 47B & \cmark & ? & \cmark \\ \hline
        7 & Mixtral-8x22B & Mistral & 141B & \cmark & ? & \cmark \\ \hline
        8 & WizardLM-2-7B & Microsoft & 7B & \cmark & ? & \xmark \\ \hline
        9 & WizardLM-2-8x22B & Microsoft & 176B & \cmark & ? & \cmark \\ \hline
        10 & Llama 3 (8B) & Meta & 8B & \cmark & 15T tokens & \xmark \\ \hline
        11 & Llama 3 (70B) & Meta & 70B & \cmark & 15T tokens & \xmark \\ \hline
        12 & DBRX & Databricks & 132B & \xmark & 12T tokens & \cmark \\ \hline
    \end{tabular}
    }
    \begin{flushleft}
    \footnotesize
    $^{\dag}$A Mixture of Experts is an architectural pattern for neural networks that splits the computation of a layer or operation (such as linear layers, MLPs, or attention projection) into multiple ``expert'' subnetworks.
    \end{flushleft}
    % \vspace{-0.8cm}

\end{table}

\subsubfour{Bias in LLMs:}
% \begin{itemize}
%     \item What metrics and how bias has been assessed in text?
%     \item Review of bias in LLMs.
% \end{itemize}
Bias in LLMs is a significant issue, which can stem from training data that can have stereotypes and prejudices embedded in them \cite{garimella_he_2021,leonardo23,schramowski_large_2022}. Recent research has heavily focused on investigating and addressing such biases in LLMs, which might be present as derogatory languages towards certain minority groups \cite{abid21,wan2023kellywarmpersonjoseph}, inconsistencies in system performance across different linguistic variations \cite{ghosh23,tarek24,treude23,patel-2021-stated}, misrepresentations of groups in society \cite{smithetal2022,Salewski23}, manifestation of historic stereotypes \cite{bartl_unmasking_2020,kotek_gender_2023,lucy-bamman-2021-gender,prakash_layered_2023,nadeem20} and general hate-filled language. 
For assessing bias in such systems, researchers have employed several methods including: 
(1) \textbf{ Embedding-based}: When testing for bias in word embeddings, word analogy tests \cite{Bolukbasi16,manzini19,papakyriakopoulos2020bias} and word association tests \cite{aylin17,pmlr-v97-brunet19a} are widely employed. 
In word analogy tests the semantic relationship between a pair of words is tested (e.g., Man : Computer programmer \(\iff\)  Woman : Homemaker). 
For word association test, bias is measured by evaluating how different classes of words (like Female names vs Male names) are associated with other words (e.g., pleasant vs unpleasant adjectives);
(2) \textbf{Template-Based}: This approach involves crafting specific templates designed to expose potential biases within language models. 
By substituting different words in placeholder positions (e.g., “The \textbf{[Name]} is a \textbf{[Occupation]}”), the model’s predictions can be analyzed for bias. For example, if ``John'' is substituted for \textbf{[Name]}, the method examines what the model predicts for [\textbf{Occupation}]~\cite{kurita-etal-2019-measuring,webster2021,ahn-oh-2021-mitigating,nadeem20,kotek_gender_2023,bartl_unmasking_2020,zhang20};
and (3) \textbf{Generated-text based}: This approach involves prompting the model and letting it generate text and then analyzing the content for biased representations. Free-text output from LLMs can be analyzed for bias using several metrics including the ones proposed by \cite{bordia_identifying_2019,liang2023, cheng_marked_2023}.

% \begin{table}[t]
% \centering
% \caption{Large Language Models (LLMs) details.}
% \label{tab:llm_details}% Preview source code for paragraph 0
% \vspace{-0.35cm}

% \resizebox{0.45\textwidth}{!}{%
% % Preview source code for paragraph 0

% \begin{tabular}{|c|c|c|c|c|c|}
% \hline 
% \textbf{Model} & \textbf{Release} & \textbf{Model} & \textbf{Pre-train} & \textbf{Open} \tabularnewline

% \textbf{} & \textbf{Year} & \textbf{Size} & \textbf{Data Scale} & \textbf{Source} \tabularnewline
% \hline
% \hline
% Databricks & 2024 & 132B & 12T & Y \\
% \hline
% GPT-3.5 Turbo & 2022 & - & - & N \\
% \hline
% GPT-4 & 2023 & - & - & N \\
% \hline
% Gemini-pro-1.5 & 2024 & - & - & N  \\
% \hline
% Gemma-7B & 2024 & 7B & 6T & Y  \\
% \hline
% Meta-Llama-3-70B & 2024 & 70B & 15T & Y \\
% \hline
% Meta-Llama-3-8B & 2024 & 8B & 15T & Y  \\
% \hline
% Mistral-7B & 2023 & 7B & - & Y \\
% \hline
% Mixtral-8x22B & 2024 & 141B & - & Y \\
% \hline
% Mixtral-8x7B & 2024 & 46.7B & -  & Y  \\
% \hline
% WizardLM-2-7B & 2024 & 7B & -  & Y \\
% \hline
% WizardLM-2-8x22B & 2024 & 141B & -  & Y \\
% \hline
% \end{tabular}
% }

% \captionsetup{justification=centering, singlelinecheck=false}
% \end{table}

\section{Dataset and LLM\lowercase{s} Overview}
% \section{Experimental Setup}
In this section, we first provide a brief description of the LLMs examined, followed by an overview of the dataset collected and utilized for our analysis.

\subsection{Large Language Models}
LLMs are advanced AI systems designed to understand, generate, and manipulate human language by leveraging vast amounts of data and sophisticated neural network architectures, often based on transformers. 
In this work, we focus on analyzing the outputs from the LLMs summarized in Table~\ref{tab:LLM-summary}.

\subsection{Data Description}
\begin{figure}[t]
\begin{centering}
\includegraphics[width = 0.65\textwidth]{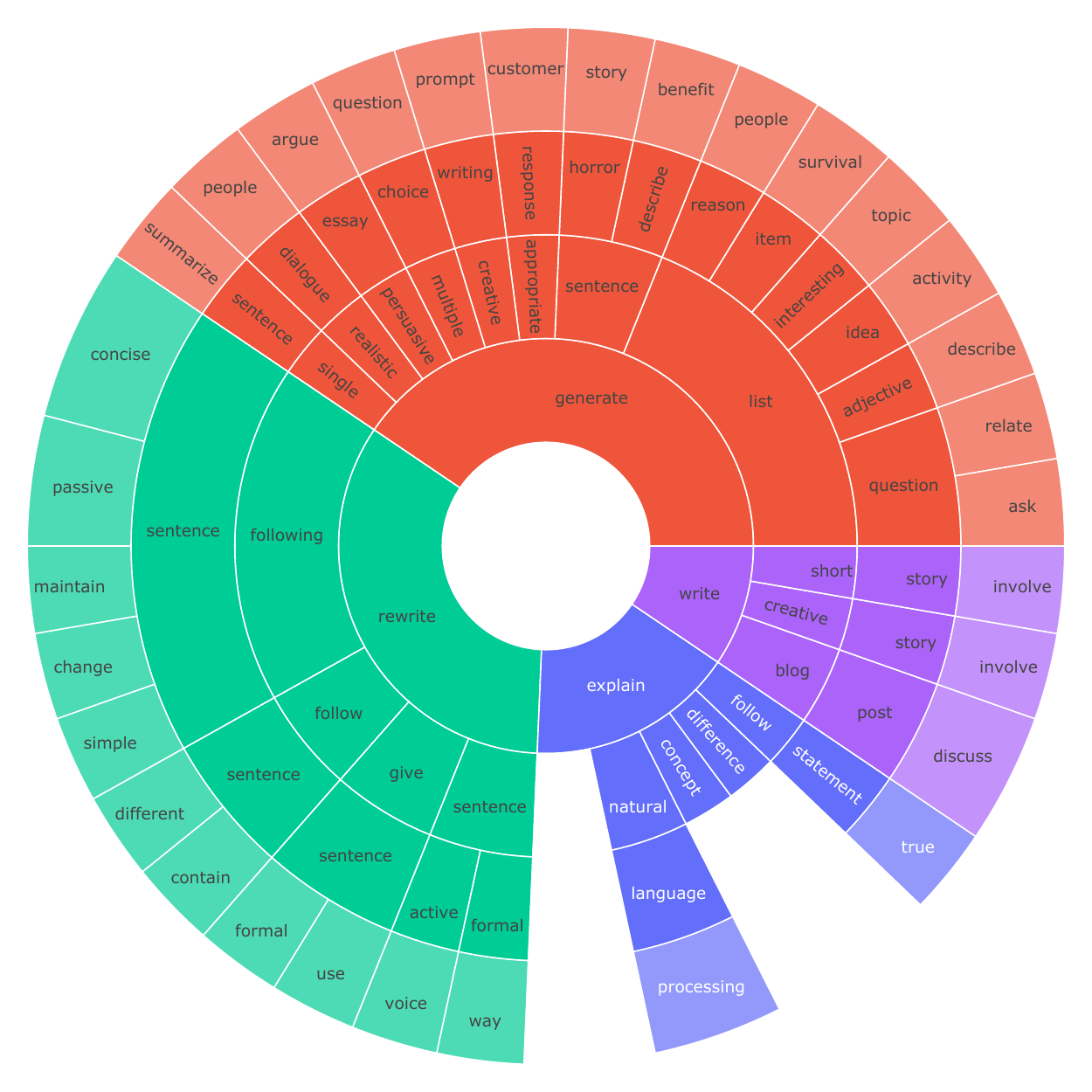}
\par
\end{centering}
\caption{Distribution of first words in prompts used to create the dataset used in this analysis.}
\label{fig:dataset_keywords}
\end{figure}

We first combined 15 prompts from the PERSUADE 2.0 corpus, introduced by Crossley et al.~\cite{crossley_2023_8221504}\footnote{\url{https://github.com/scrosseye/persuade_corpus_2.0}}, with a selection of prompts/instructions from the Stanford Alpaca project dataset~\cite{alpaca}\footnote{\url{https://github.com/tatsu-lab/stanford_alpaca}}. 
The PERSUADE 2.0 dataset was chosen for its human-generated text corresponding to each prompt/instruction, while the Stanford Alpaca dataset was included for its rich number of diverse prompts.
This combination resulted in a comprehensive set of 5,015 unique prompts, predominantly featuring keywords depicted in Figure~\ref{fig:dataset_keywords}. 
These instructions spanned a variety of topics, from learning artificial intelligence to generating persuasive essays.
Finally, each prompt was used to generate 50 texts from each LLM, resulting in a total of approximately 3M texts, including those written by humans.

% Recognizing the need for a more focused dataset to accurately analyse the outputs of each LLM, we curated our initial, broader dataset. We selectively retained instructions that specifically focused on narrative tasks, such as story generation and textual rewriting. 
% This curation effectively narrowed our dataset with the overall details provided in Table~\ref{tab:dataset_statistics}. 

% The finalized dataset utilized in our research comprised 1,505 unique instructions, predominantly featuring keywords depicted in Figure~\ref{fig:dataset_keywords}. These instructions spanned a variety of topics, from learning artificial intelligence to generating persuasive essays, with the distribution across 26 distinct topics illustrated in Figure~\ref{fig:dataset_topics}. These topic clusters were extracted using BERTopic\footnote{\url{https://github.com/MaartenGr/BERTopic}} ~\cite{grootendorst2022bertopic}.

% \begin{figure}[t]
% \begin{centering}
% \includegraphics[width = 0.5\textwidth]{images/topics_and_keywords/topics.png}
% \par
% \end{centering}
% \caption{Dataset keywords.}
% \label{fig:dataset_topics}
% \end{figure}

Detailed statistics of the resulted dataset are provided in Table~\ref{tab:dataset_statistics}. 
For example, for the given prompts, the average word count is 202, the average sentence count is 13.
Also, out of the 250,750 texts generated using GPT-3.5 for all prompts, there are 85\% unique words, entropy ratio of 14, and a lexical diversity of 3.5.

Finally, Table~\ref{tab:model_hyperparameters} presents the generation hyperparameters used across all  LLMs evaluated in this study.
These parameters—including maximum tokens, temperature, top-p, frequency penalty, and repetition penalty—play a critical role in shaping the style, diversity, and consistency of the outputs produced by each model.

% \textcolor{red}{At first glance, we note that there is a clear discrepancy in the vocabularies (number of unique words) among the models. 
% For instance, despite GPT-4 generating fewer texts compared to Gemini-pro, its vocabulary is substantially larger (21k vs. 14k).
% Additionally, it's noteworthy that GPT-4 exhibits superior vocabulary density, showcasing a broader lexicon in its language generation. 
% Finally, despite GPT-3.5 and humans having similar densities, the richness of vocabulary in GPT-3.5 remains comparable to that of humans, even with significantly less text.}

\begin{table}[t]
\centering
\caption{Dataset details and statistics.}
\label{tab:dataset_statistics}% Preview source code for paragraph 0
% \vspace{-0.35cm}

\footnotesize{Prompts Summary} \\
\vspace{0.1cm}

\resizebox{0.45\textwidth}{!}{%
% Preview source code for paragraph 0

\begin{tabular}{|c|c|c|c|c|}
\hline 
\textbf{\#Prompts} & \textbf{\#Unique } & \textbf{Average} & \textbf{Average} & \textbf{\#topics} \tabularnewline
\textbf{} & \textbf{Words} & \textbf{Words} & \textbf{Sentences} & \textbf{} \tabularnewline
\hline 
\hline 
5,015 & 96,606 & 202 & 13 & 26 \tabularnewline
\hline 
\end{tabular}
}

\vspace{0.1cm}
\footnotesize{Text Measures} \\
\vspace{0.1cm}
\resizebox{0.65\textwidth}{!}{%
% Preview source code for paragraph 0

\begin{tabular}{|c|c|c|c|c|c|c|}
\hline 
\textbf{Model} & \textbf{\#texts}& \textbf{Unique Word} & \textbf{Entropy} & \textbf{Monosyllable} & \textbf{Polysyllable} & \textbf{Lexical} \tabularnewline

\textbf{} & \textbf{} & \textbf{Ratio} &\textbf{Ratio} & \textbf{Ratio} & \textbf{Ratio} & \textbf{Diversity} \tabularnewline
\hline
\hline
Databricks&250,750 & 75.0 & 10.8 & 64.3 & 14.3 & 3.5 \\
\hline
GPT-3.5 Turbo&250,750 & 85.0 & 17.8 & 67.2 & 12.7 & 5.9 \\
\hline
GPT-4&250,750 & 86.4 & 18.1 & 66.4 & 13.3 & 6.1 \\
\hline
Gemini-pro-1.5&250,750 & 69.5 & 11.9 & 62.0 & 16.0 & 5.3 \\
\hline
Gemma-7B&250,750 & 71.8 & 11.2 & 60.6 & 16.6 & 3.0 \\
\hline
Meta-Llama-3-70B&250,750 & 60.6 & 6.6 & 67.0 & 12.9 & 1.4 \\
\hline
Meta-Llama-3-8B&250,750 & 61.2 & 7.0 & 67.7 & 12.5 & 1.5 \\
\hline
Mistral-7B&250,750 & 64.9 & 8.7 & 66.7 & 13.1 & 2.0 \\
\hline
Mixtral-8x22B&250,750 & 66.1 & 8.9 & 65.9 & 13.6 & 2.3 \\
\hline
Mixtral-8x7B&250,750 & 62.6 & 7.1 & 67.4 & 12.7 & 1.3 \\
\hline
WizardLM-2-7B&250,750 & 64.3 & 7.8 & 65.4 & 14.1 & 2.1 \\
\hline
WizardLM-2-8x22B&250,750  & 64.3 & 7.9 & 65.3 & 13.9 & 2.2 \\
\hline
\hline
Human& 25,000& 43.0 & 1.9 & 78.9 & 5.2 & 0.1 \\
\hline
\end{tabular}
}

\captionsetup{justification=centering, singlelinecheck=false}
\end{table}

\begin{table}[t]
\centering
\caption{Model generation hyperparameters.}
\label{tab:model_hyperparameters}
\resizebox{0.8\textwidth}{!}{%
\begin{tabular}{|c|c|c|c|c|c|}
\hline 
\textbf{Model} & \textbf{Max Tokens}& \textbf{Temperature} & \textbf{Top-p} & \textbf{Frequency Penalty} & \textbf{Repetition Penalty} \\
\hline 
\hline
Databricks & 512 & 0.7 & 0.9 & 0.0 & 1.0 \\
\hline
GPT-3.5 Turbo & - & 0.7 & 1.0 & 0.0 & 1.0 \\
\hline
GPT-4 & 250 & 0.7 & 1.0 & 0.0 & 1.0 \\
\hline
Gemini Pro 1.5 & - & 1.0 & 0.95 & - & 1.0 \\
\hline
Gemma-7B & 512 & 1.0 & 1.0 & 0.0 & 1.0 \\
\hline
Meta-Llama-3-70B & 512 & 0.7 & 0.9 & 0.0 & 1.0 \\
\hline
Meta-Llama-3-8B & 512 & 0.7 & 0.9 & 0.0 & 1.0 \\
\hline
Mistral-7B & 512 & 0.7 & 0.9 & 0.0 & 1.0 \\
\hline
Mixtral-8x22B & 512 & 0.7 & 0.9 & 0.0 & 1.0 \\
\hline
Mixtral-8x7B & 512 & 0.7 & 0.9 & 0.0 & 1.0 \\
\hline
WizardLM-2-7B & 512 & 0.7 & 0.0 & 0.0 & 1.0 \\
\hline
WizardLM-2-8x22B & 512 & 0.7 & 0.9 & 0.0 & 1.0 \\
\hline
\end{tabular}
}
\end{table}

\section{Empirical Evaluation}
In this section, we conduct a series of evaluations and discussions aimed at thoroughly examining the analyzed LLMs, with the intent of answering the research questions outlined above.

% \begin{figure}[t]
% \begin{centering}
% \subfloat[\footnotesize{Inner-similarity between language models using edit distance.}]{\includegraphics[width = 0.45\textwidth]{images/word-level-edit-distance-similarity/inner_edit_similarity_box_plot.pdf}} \hfill
% \subfloat[\footnotesize{Inner-similarity between language models using cosine similarity.}]{\includegraphics[width = 0.45\textwidth]{images/word-level-edit-distance-similarity/inner_cos_similarity_box_plot.pdf}}
% \par
% \end{centering}
% \caption{Comparison of inner-similarity between language models.}
% \label{fig:inner-similarity-comparison}
% \end{figure}

% \begin{figure}[t]
% \begin{centering}
% \subfloat[\footnotesize{Inter-similarity using edit distance.}]{\includegraphics[width = 0.45\textwidth]{images/word-level-edit-distance-similarity/Human_similarity_box_plot.pdf}} \hfill
% \subfloat[\footnotesize{Inter-similarity using cosine similarity.}]{\includegraphics[width = 0.45\textwidth]{images/word-level-edit-distance-similarity/Human__cosine_similarity_box_plot.pdf}}
% \par
% \end{centering}
% \caption{Comparison of inter-similarity against human-written text.}
% \label{fig:human-inter-similarity-comparison}
% \end{figure}

% \begin{figure}[t]
% \begin{centering}
% \includegraphics[width = 0.43\textwidth]{images/word-level-edit-distance-similarity/inner_similarity_source_comparisons_plot.pdf}
% \par
% \end{centering}
% \caption{Inner-similarity between proprietary and non-proprietary.}
% \label{fig:inner-word-level-edit-distance-by-types}
% \end{figure}

\subsection{RQ1: Comparative Analysis of LLM Texts } 
To address \greyball{RQ1}, we conduct a text similarity analysis and apply various readability statistics to assess the data, as detailed below.

\subsubsection{Text Similarity Analysis } 
For analyzing the similarity between the outputs of the examined LLMs, the procedure is as follows: 
For each prompt, we compare each generated text to others associated with the same prompt. 
Specifically, we compute pairwise similarities using both \textit{cosine similarity} and \textit{Word-Level Levenshtein Edit Distance}, which measures the number of single-word edits required to transform one text into another.
This comparison is conducted across various contexts, including within the same LLM, among different LLMs, or with human-generated texts.
This approach allows for a comprehensive exploration of text similarities across different scenarios and models.

\begin{figure}[t]
\begin{centering}
\includegraphics[width = 0.55\textwidth]{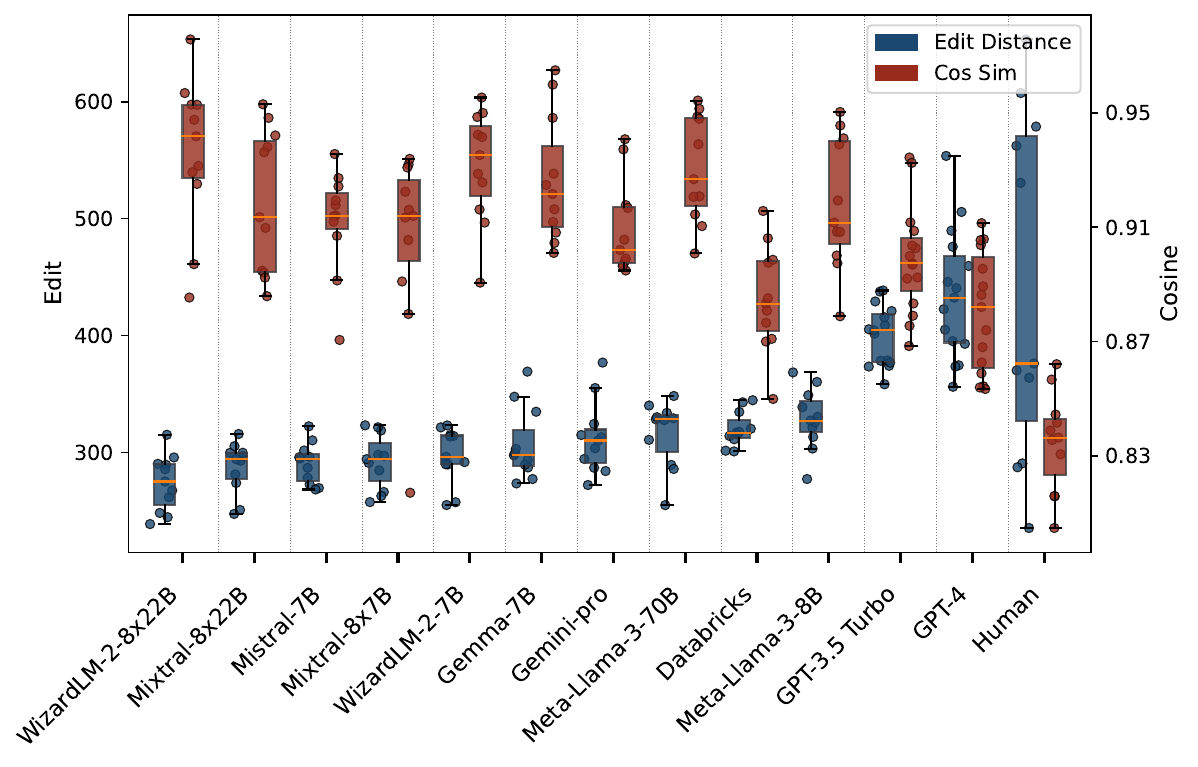}
% \hfill
% \subfloat[\footnotesize{Inner-similarity  using cosine similarity.}]{\includegraphics[width = 0.45\textwidth]{images/word-level-edit-distance-similarity/inner_cos_similarity_box_plot.pdf}}
\par
\end{centering}
% \vspace{-0.5cm}
\caption{Comparison of inner-text similarity.}
\label{fig:inner-similarity-comparison}
\end{figure}

\subsubfour{Inner-text similarity:} 
The results of this analysis are presented in Figure~\ref{fig:inner-similarity-comparison}, from which we make the following notes:
(i) Humans exhibit lower word-level similarity to one another compared to LLMs, reflecting the unique and individualistic nature of human writing styles. 
Human written text also show higher variance in inner-similarity, while some LLMs tend to show more consistent similarity levels.
(ii) Among LLMs, WizardLM-2-8x22b shows the highest similarity in generated text, followed by Llama-3-70b and WizardLM-2-7b.
(iii) Finally, GPT-4 has the lowest similarity to its own outputs, aligning with the stylometric features seen in Table~\ref{tab:dataset_statistics}, indicating GPT-4's high lexical diversity and unique word ratio compared to other LLMs.
In summary, these findings collectively indicate a degree of unpredictability in the outputs of LLMs.

\begin{figure}[h]
\begin{centering}
\includegraphics[width = 0.65\textwidth]{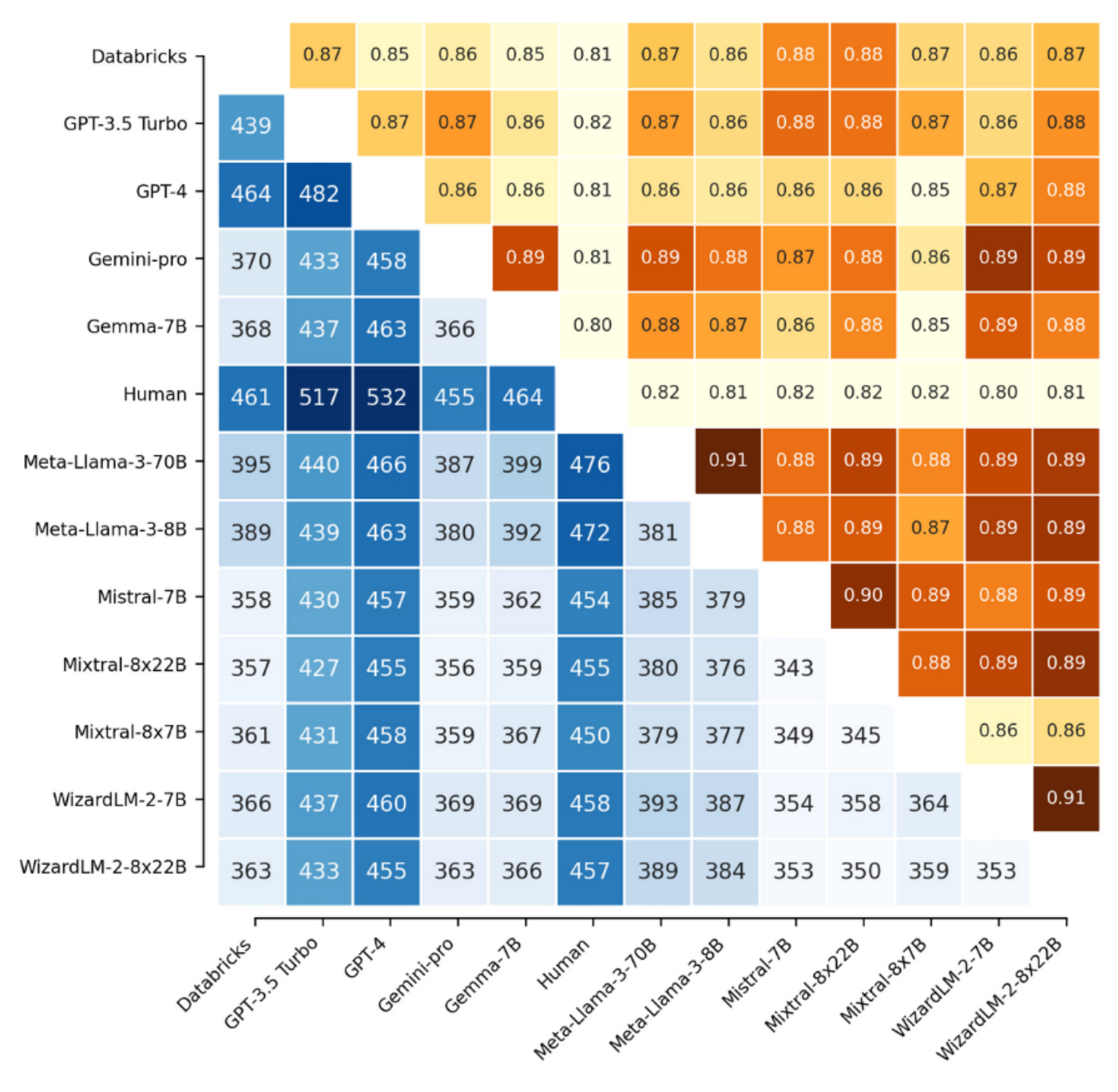}
\par
\end{centering}
% \vspace{-0.43cm}
\caption{Comparaison of (average) inter-text similarity. 
The lower part (blue) displays the similarity calculated using word-level edit distance, while the upper part (orange) illustrates the similarity determined by cosine similarity.
}
\label{fig:overall-word-level-edit-distance}
\end{figure}

\subsubfour{Inter-text similarity:}
The findings from this analysis are depicted in Figure~\ref{fig:overall-word-level-edit-distance}, leading to the following observations:
(i) Human-written text shows the least similarity to all LLMs. Among the LLMs, Mistral appears to be the most similar to human-written text, while OpenAI models are the least similar.
(ii) There is a notable similarity between the Llama 3 models and Mistral, as well as with WizardLM-2 models. Llama-3-8B and 70B exhibit the highest similarity among all LLMs, whereas GPT-4 and GPT-3.5 display the least similarity, highlighting the diversity between these two models.
(iii) Lastly, while GPT-3.5 shares some similarity with models like Llama 3 and Mixtral-8xx22B, GPT-4, similar to human-written text, shows no strong similarities with any other models. Among all LLMs, GPT-4 is the least similar to others, with human-written text being the least similar overall.
These findings collectively underscore the nuanced patterns in text similarity across different models and underscore the complexity of language model outputs.

% The findings from this analysis are depicted in Figure~\ref{fig:overall-word-level-edit-distance} and Figure~\ref{fig:inter-word-level-edit-distance-by-types}, leading to the following observations:

% (1) First, as anticipated, we can notice that human-written text shows the least similarity to all LLMs. However, when it comes to the most similar among all LLMs, it appears to be Mistral, and the least similar to OpenAI models.

% (2) Second, there is an obvious similarity between the Llama 3 models and Mistral, as well as with WizardLM-2 models. Llama-3-8B and 70B share the highest similarity among all LLMs, with GPT-4 and GPT-3.5 sharing the least in our findings. This points out the diversity between the two models.

% (3) While GPT-3.5 Turbo shares some similarity to models such as both Llama 3 and Mixtral-8xx22B, GPT-4, similar to human-written text, shares no strong obvious similarities. Out of all LLMs, GPT-4 is the least similar to all other LLMs, with human-written text being the least similar.

\subsubsection{Stylometric Analysis}
In this section, we utilize six readability statistics to evaluate the text data. 
Specifically, we:
(1) calculate the number of words in a text that are considered difficult to read or understand, defined as those not in a predefined list of common, easy-to-understand words or contain more than two syllables;
(2) apply the Flesch Reading Ease score to assess the overall readability of the text~\cite{flesch1948new};
(3) estimate the time required for an average reader to read the text, based on word count and standard reading speed~\cite{demberg2008data}; 
(4) determine the estimated reading level or grade level of the text by aggregating results from various readability tests, indicating the educational grade level necessary to comprehend the content, such as ``5th grade'' or ``college level''; and
(5) calculate number of monosyllabic words and (6) polysyllabic words.

\begin{figure*}[h]
\begin{centering}

\subfloat[Difficult words.]{\includegraphics[width = 0.3\textwidth]{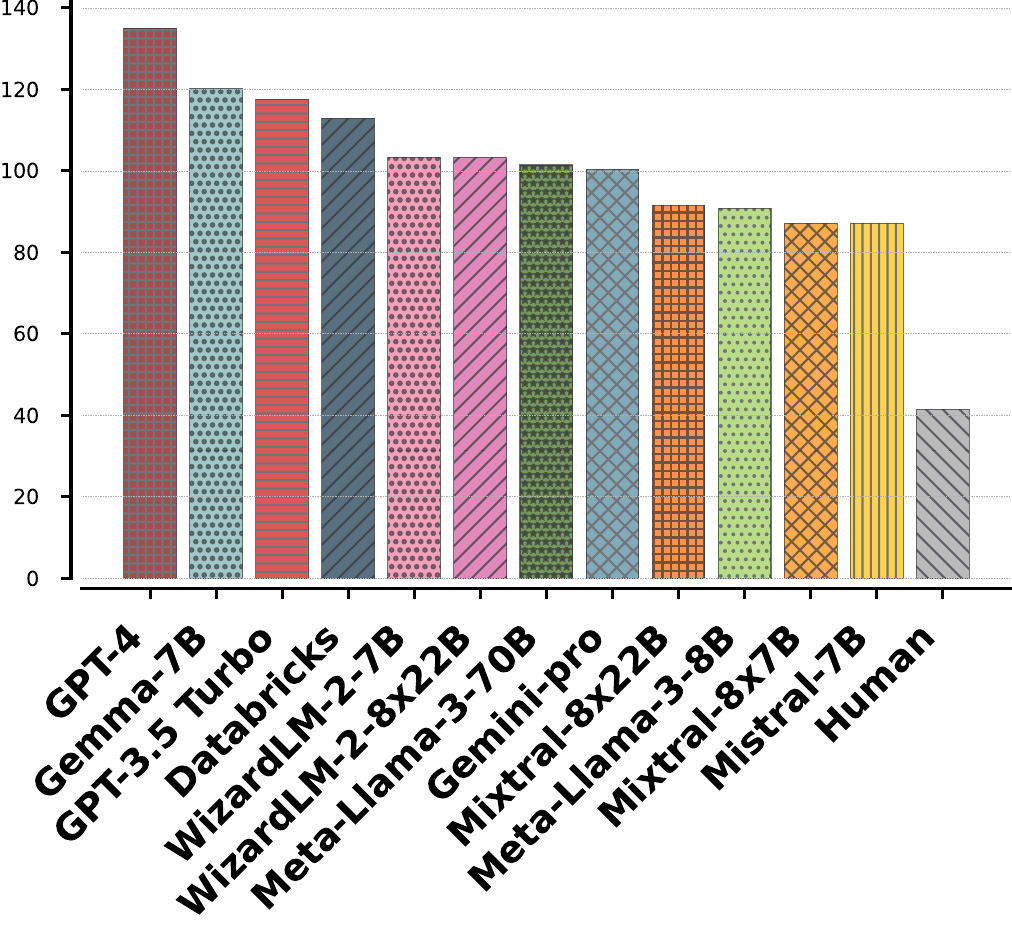}
}
\subfloat[Flesch reading ease.]{\includegraphics[width = 0.3\textwidth]{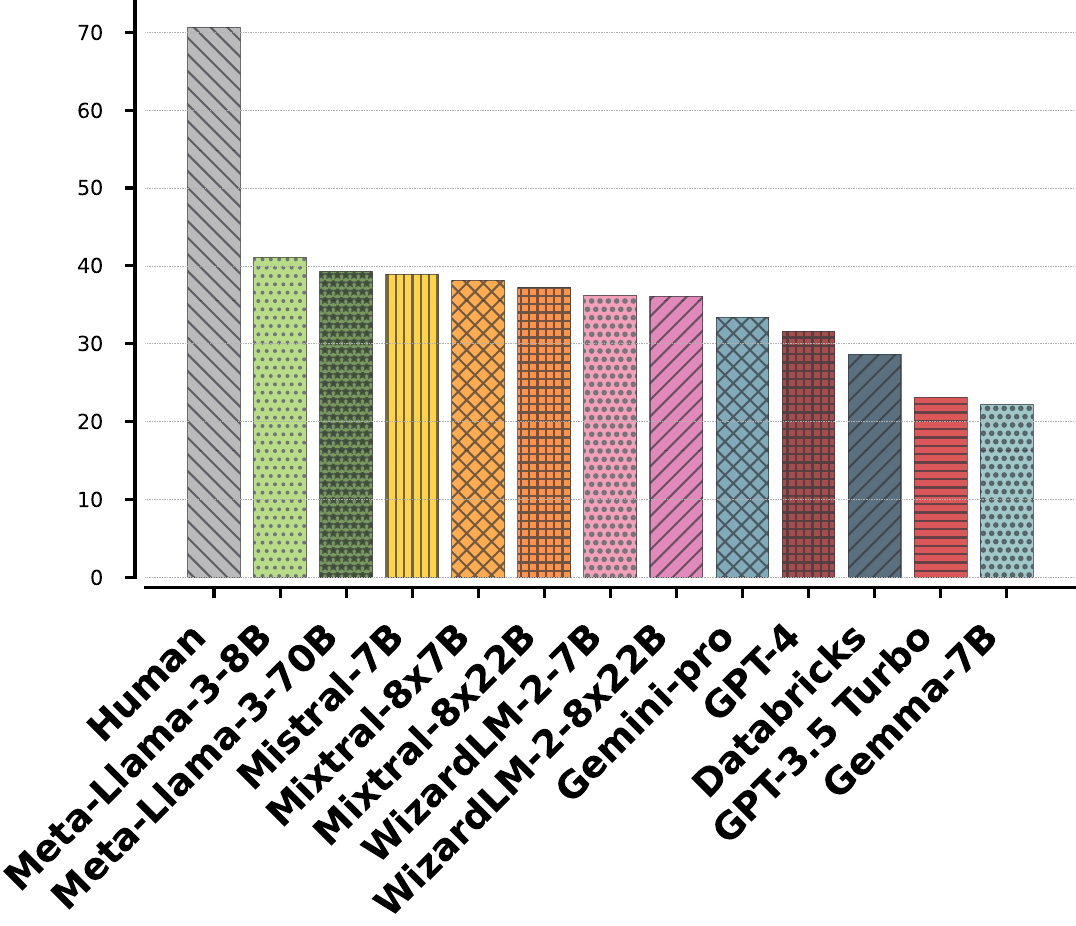}
\label{fig:readability_num_years}
}
\subfloat[Reading time.]{\includegraphics[width = 0.3\textwidth]{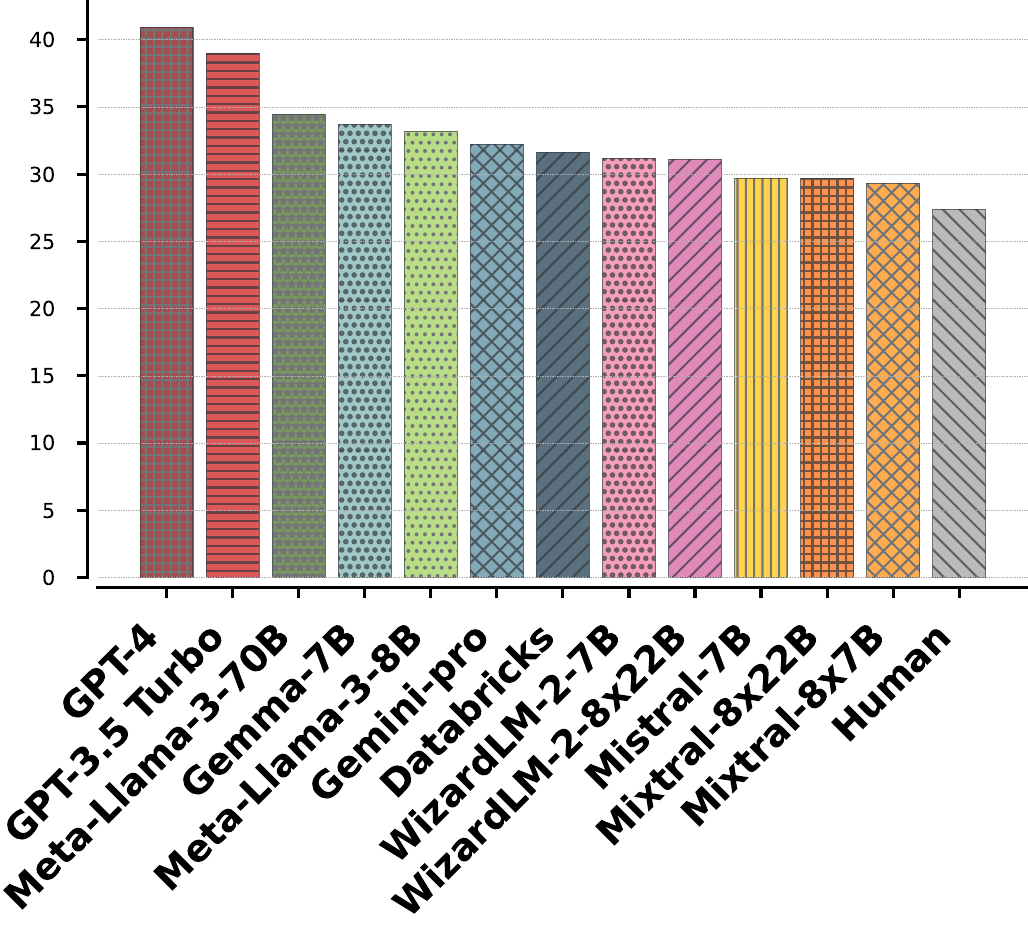}
}\\
\subfloat[Text standard.]{\includegraphics[width = 0.3\textwidth]{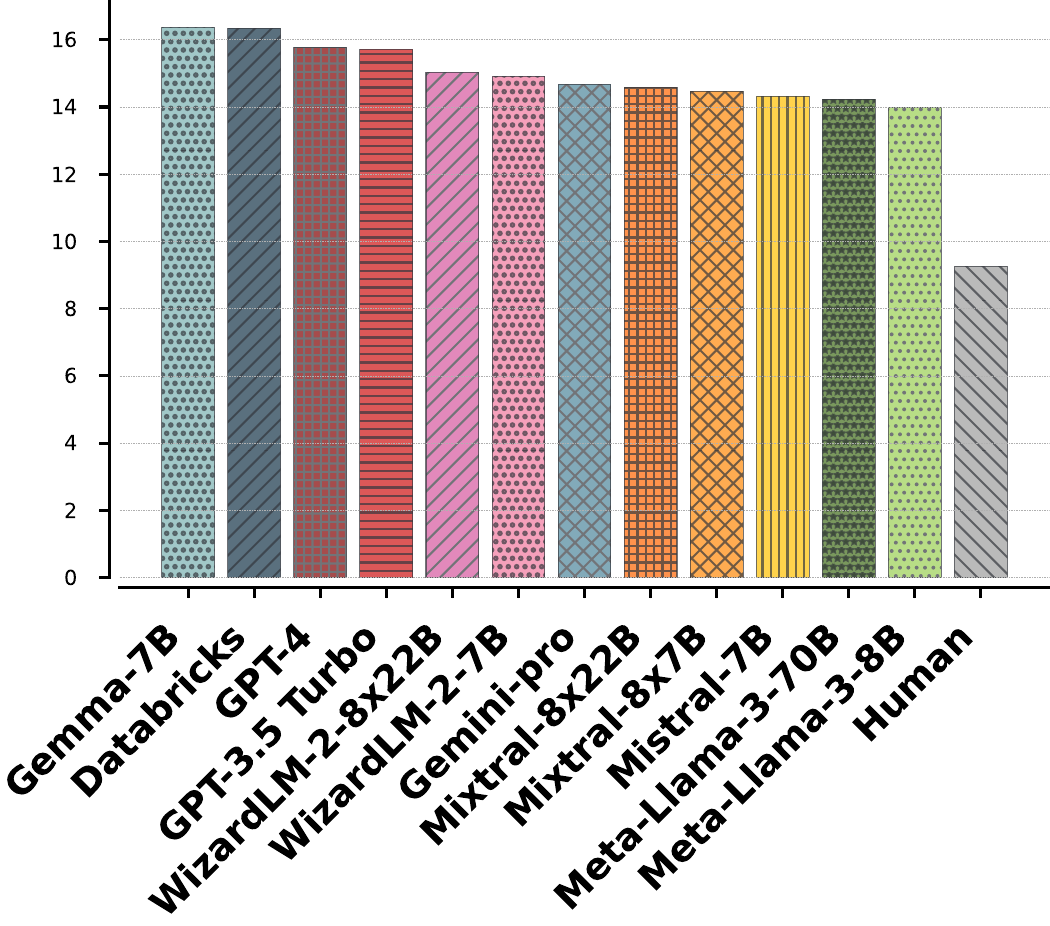}
}
\subfloat[Monosyllabic words.]{\includegraphics[width = 0.3\textwidth]{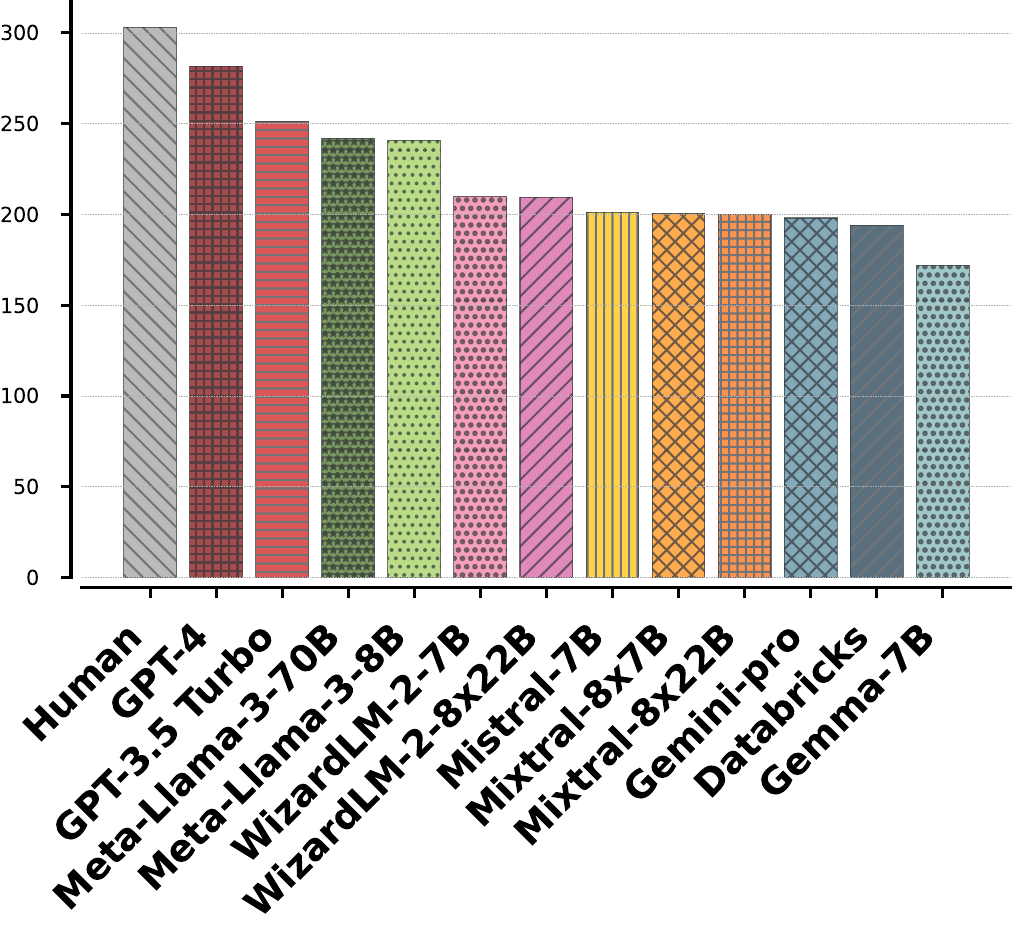}
}
\subfloat[Polysyllabic words.]{\includegraphics[width = 0.3\textwidth]{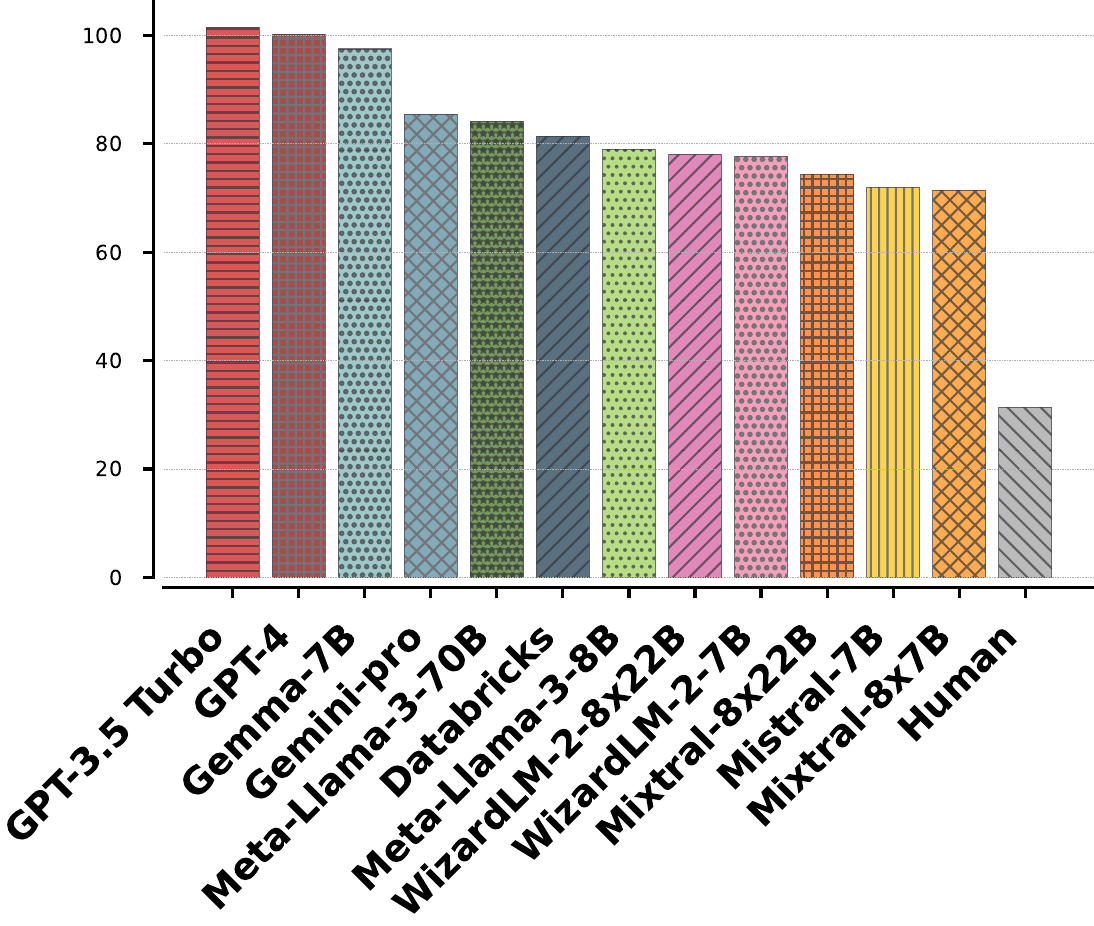}
}
\par
\end{centering}
\caption{Readability Statistics.}
\label{fig:readability-statistics}
\end{figure*}

The obtained results are illustrated in Figure~\ref{fig:readability-statistics}.
% Also, we explored various readability stylometric features, which reveals distinct differences in linguistic styles between human and LLM-generated texts.
Notably, when comparing the writing of LLMs to that of humans, it becomes apparent that human writing tends to be simpler, more accessible, and easier to read. 
Humans often use straightforward language and balanced punctuation. 
In contrast, models like GPT-4 and GPT-3.5 produce more complex and richly detailed content, characterized by a higher frequency of challenging vocabulary.
% \subsubsection{Readability Measures:} detailed are provided in Table~\ref{tab:read-measures} where we found the following:
% (1) Initially, we assessed the density of vocabulary within the texts by examining the frequency of words with three or more syllables. A higher count indicates a richer lexicon, with GPT-4 showing the highest ratio of such words, implying its preference for utilizing more complex language. We found that human-generated text features the lowest count, suggesting a inclination for simpler and more readily understandable language.
Subsequently, the Flesch Reading Ease score, used to assess textual readability~\cite{flesch1948new}, indicates that human-written content is the most readable, achieving the highest score. 
In contrast, Gemma-7B received the lowest score, reflecting its complexity and reduced accessibility.
ALso, reading time analysis~\cite{demberg2008data} reveals that GPT-4 requires the longest reading duration, likely due to its lengthier and potentially more complex style, while human-generated text demands the shortest reading time.
Finally, the complexity in LLM writing is
further substantiated by their use of polysyllabic words, in contrast to humans who often prefer monosyllabic words for simplicity.

\begin{tcolorbox}[colback=gray!10, colframe=black, boxsep=0.01mm, left=1mm, right=1mm, title=Summary of Key Findings for \greyball{RQ1}]
This analysis reveals that human-written texts show greater variability and lower similarity compared to LLM-generated texts, emphasizing the unique nature of human writing. 
Among LLMs, WizardLM-2-8x22b produced the most consistent outputs, while GPT-4 exhibited the highest lexical diversity, resulting in the lowest similarity to its own outputs. 
In inter-text comparisons, human-written content was the least similar to LLM outputs, with Mistral being closest to human text and GPT-4 showing the greatest divergence from other models. 
Readability analysis further highlighted that human texts are generally simpler and more accessible, while LLM outputs, especially from GPT-4, are more complex and challenging.
\vspace{-0.25cm}
\end{tcolorbox}

\begin{figure*}[t]
\begin{centering}

\subfloat[Prompt 1.]{\includegraphics[width = 0.3\textwidth]{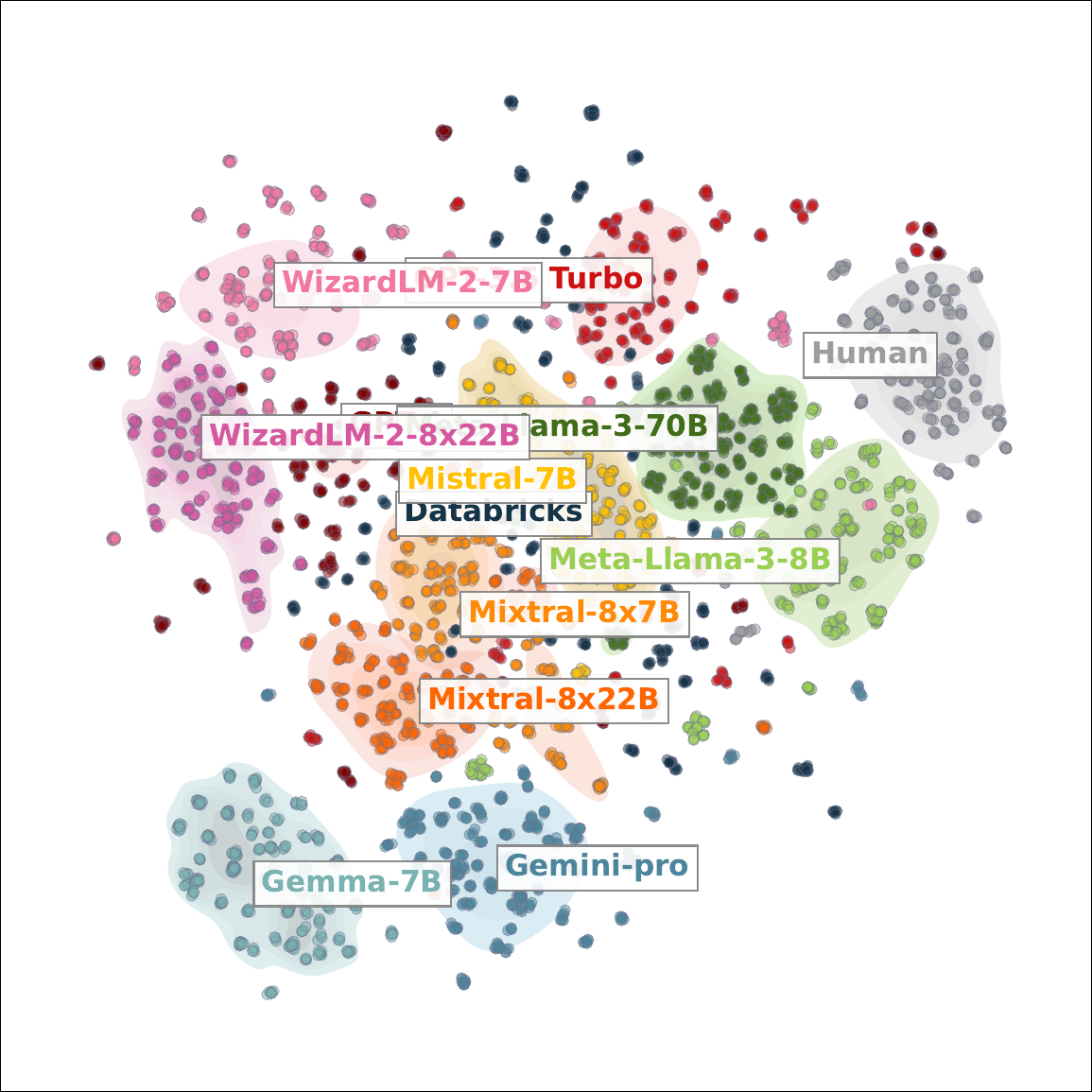}
}
\subfloat[Prompt 2.]{\includegraphics[width = 0.3\textwidth]{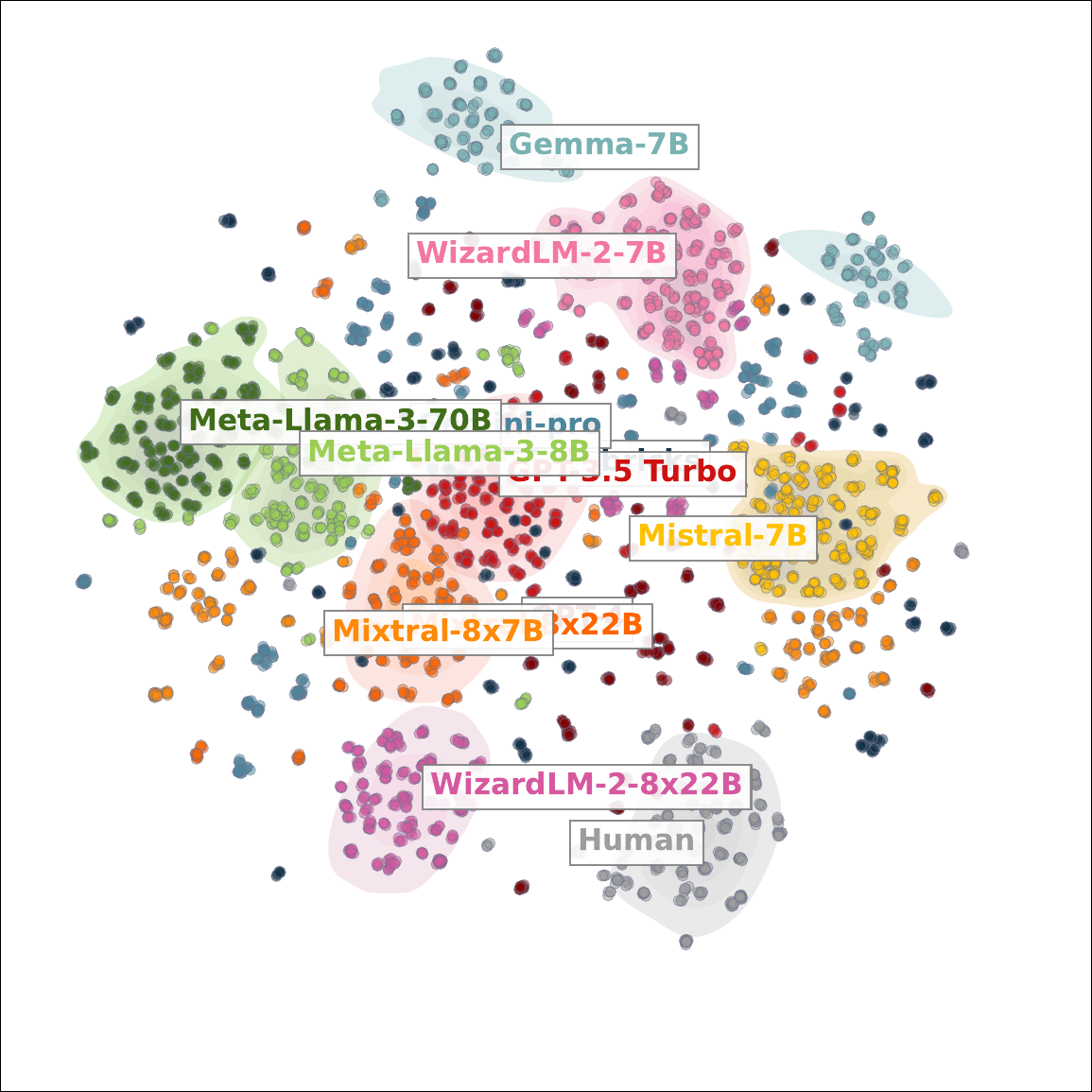}
}
\subfloat[Prompt 3.]{\includegraphics[width = 0.3\textwidth]{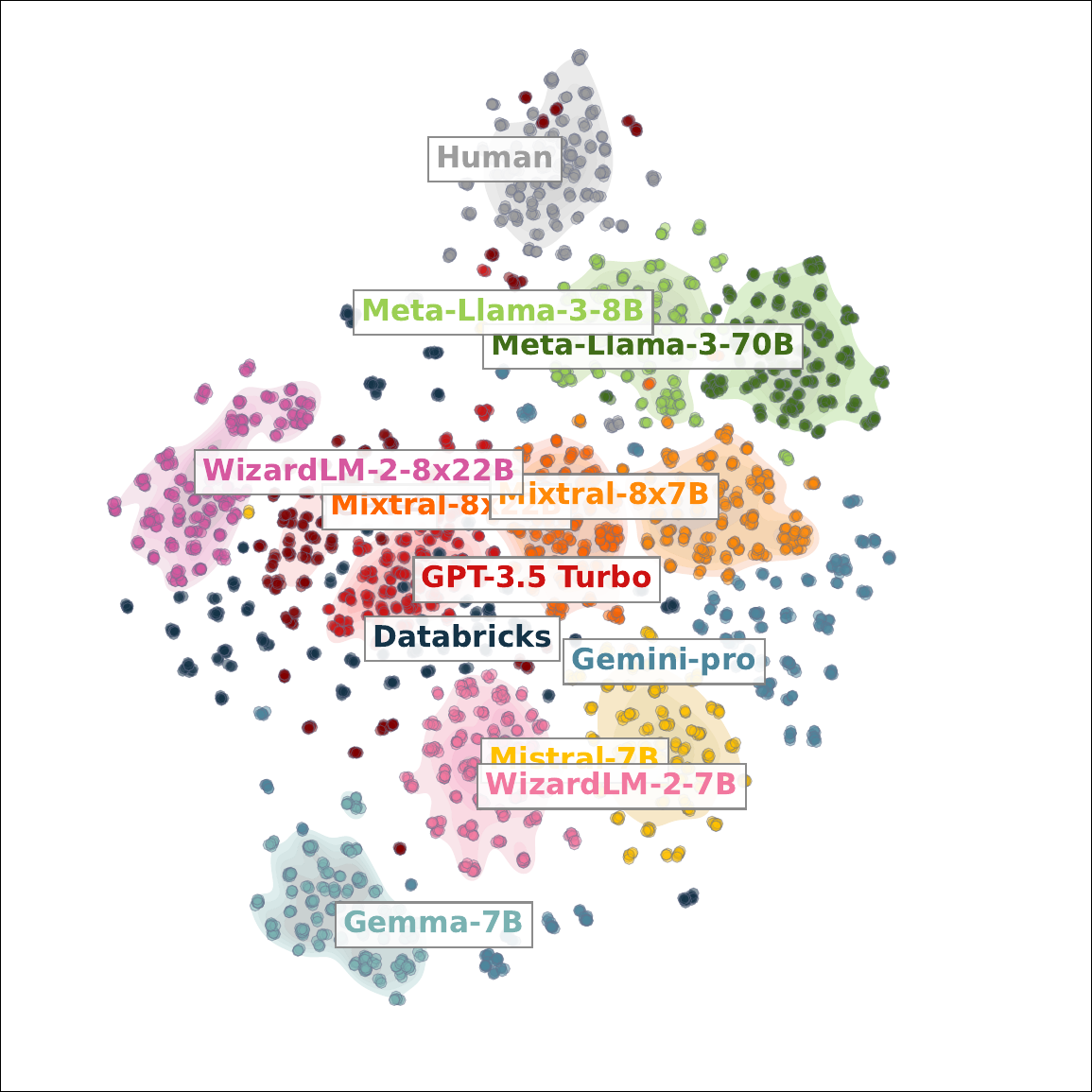}
}\\
\subfloat[Prompt 4.]{\includegraphics[width = 0.3\textwidth]{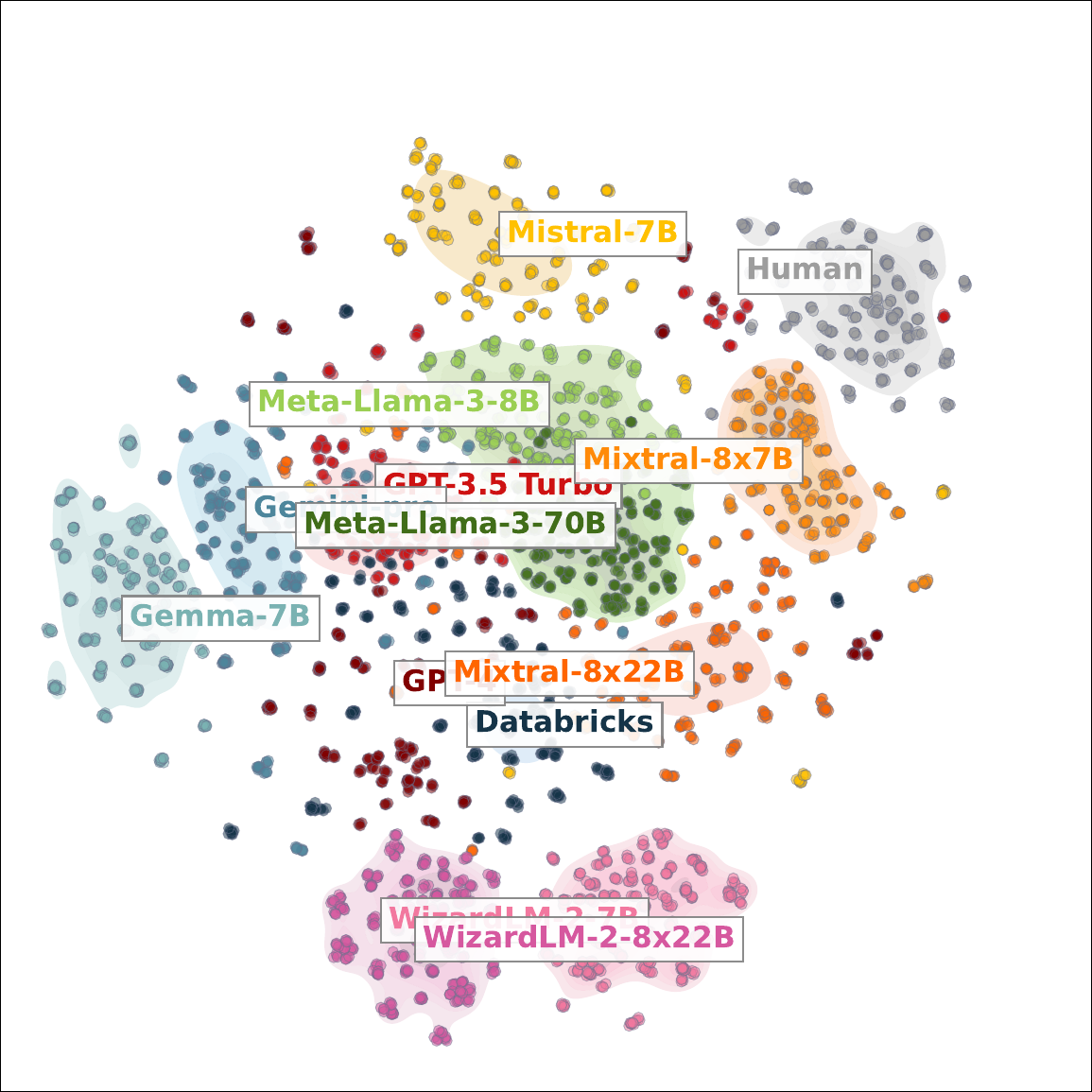}
}
\subfloat[Prompt 5.]{\includegraphics[width = 0.3\textwidth]{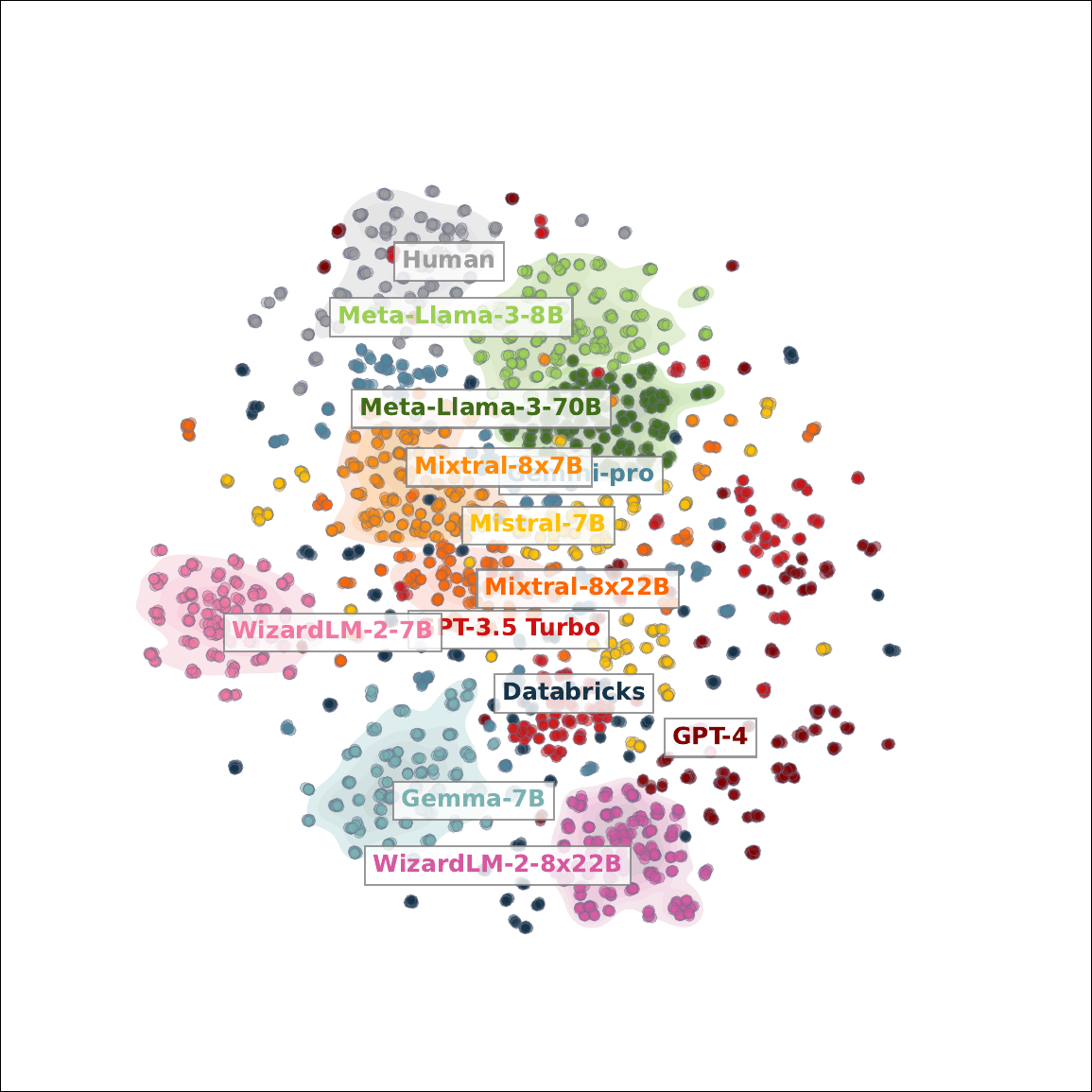}
}
\subfloat[Prompt 6.]{\includegraphics[width = 0.3\textwidth]{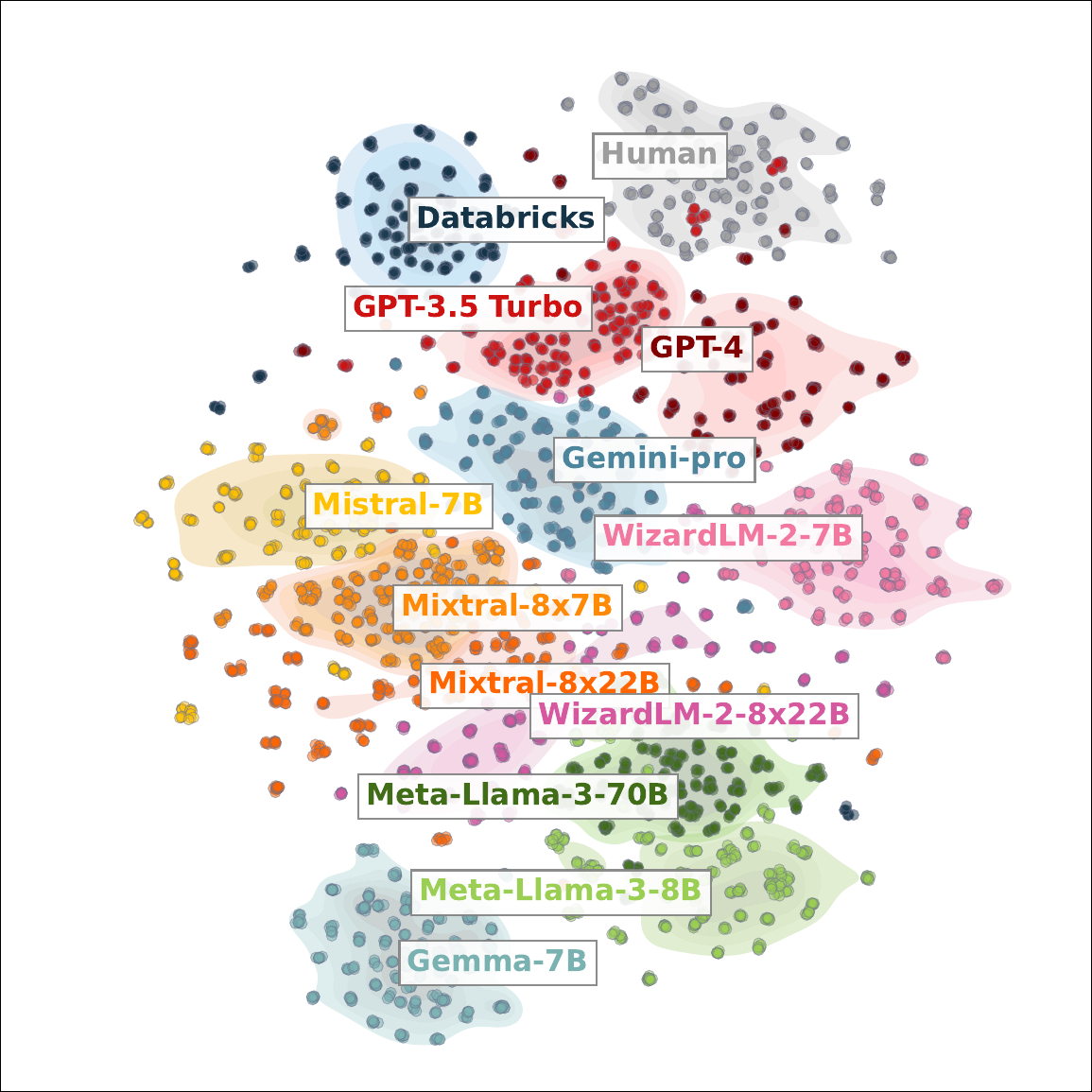}
}
\par
\end{centering}
\caption{UMAP projection of high-dimensional text generated into a two-dimensional space for visualization.}
\label{fig:UMAP_Viz}
\end{figure*}

% \subsection{Similarity clustering}

% The findings from this analysis are shown in Figure~\ref{fig:UMAP_Viz}, where we examine the embeddings generated using a bag-of-words approach and cosine similarity. To focus on writing style rather than topic clustering, we analyzed texts per instruction rather than clustering all instructions together. We utilized Uniform Manifold Approximation and Projection (UMAP) to reduce the dimensionality of the sparse matrix to two dimensional and set the metric parameter to cosine similarity and increased size of $n$ to 100 to better capture the global structure of the data. This approach provided the following insights:

% (i) There are obvious clusters between all groups, with Gemma-7B and human-written text showing to have the less similarity to the other language models.

% (ii) Certain models consistently show high similarity, such as the Llama 3 models and Mistral series.

% (iii) GPT-4 has the most spread in comparison to all other language models. This is consistent throughout all four prompts.

% These insights highlight the variability and similarity in text generation patterns among different models and human-written texts.

\subsection{RQ2: Variance in Text Generation}

The inner similarity analysis presented in Figure~\ref{fig:inner-similarity-comparison} reveals significant variance in text generation among different LLMs.
To illustrate this, Figure~\ref{fig:UMAP_Viz} employs the UMAP algorithm~\cite{McInnes2018} to visualize high-dimensional text data in a comprehensible two-dimensional space, for six different prompts.
This visualization offers a clear representation of the distribution and relationships among the text outputs, facilitating the identification of patterns and distinctions across the diverse LLMs under examination.

At first glance, we note that some LLMs display notable variability in text generation, whereas others exhibit a more uniform and consistent output.
For instance, it is observed that Mistral 7B demonstrates low variance in text generation, while Gemini-pro-1.5 exhibits higher variance. 
This disparity suggests that Mistral 7B tends to produce more consistent and uniform text outputs, while Gemini-pro-1.5 introduces greater variability in its generated text.
This insight into the diversity of LLM behavior is crucial for understanding the nuances of these models.

\begin{tcolorbox}[colback=gray!10, colframe=black, boxsep=0.01mm, left=1mm, right=1mm, title=Summary of Key Findings for \greyball{RQ2}]
This analysis reveals that some models produce more consistent and uniform outputs, while others show greater variability, highlighting the diversity in LLM behavior.
\end{tcolorbox}

\subsection{RQ3: Classification Performance}

The ability to differentiate between text written by humans and that generated by LLMs holds significant importance in various contexts. 
In applications such as content moderation, misinformation detection, and ensuring ethical use of AI, being able to identify the source of text content is critical. Furthermore, understanding which specific LLM has generated a text can provide valuable insights into the model's biases, tendencies, and potential shortcomings.
This capability not only enhances transparency and accountability in AI applications but also aids in addressing ethical concerns surrounding the use of language models. Moreover, it equips users, researchers, and policymakers with the necessary tools to assess the reliability and trustworthiness of textual content, thereby encouraging a more responsible and conscious integration of LLM-generated content across various sectors.

To achieve this objective, we make use of BERT~\cite{devlin-etal-2019-bert}, several variants of DeBERTa-v3~\cite{he2023debertav3} (DeBERTa-v3-xsmall, DeBERTa-v3-small, DeBERTa-v3-base), and an XGBoost~\cite{Chen_2016} model with Bag of Words features. 
We partition our dataset into TrainVal-Test sets using a split of 60\%, 20\%, and 20\%, respectively.
% We trained language models such as BERT and DeBERTa for one epoch and used XGBoost out of the box. The parameters used for training included a learning rate of 1e-5, a maximum length of 512, batch-size of 4, weight decay at 0.01 and with no warm up ratio.
The performance of these classifiers is detailed in Table~\ref{tab:classification_performance}, which presents the achieved classification metrics and reveals the following insights:
\textbf{BERT} outperforms all DeBERTa variants as well as XGBoost with Bag-of-Word (BoW). BERT achieves the highest overall accuracy of \textbf{0.7095} and demonstrates superior performance across multiple LLMs, with the highest F1-scores in most categories. This highlights its robustness in identifying text generated by different models.
\textbf{DeBERTa-v3-base} shows strong performance, particularly by achieving the highest F1-score (\textbf{0.7231}) for the DBRX class. Among the DeBERTa variants, it consistently outperforms the smaller models, demonstrating the effectiveness of a larger model architecture in learning various stylometric patterns.
\textbf{DeBERTa-v3-small} and \textbf{DeBERTa-v3-xsmall} provide reasonable performance but lag behind BERT and the larger DeBERTa model. Finally, \textbf{XGBoost-BoW} exhibits the highest F1-score (\textbf{0.9906}) for the Human class, indicating its strength in fitting strongly to human-written text. However, it generally performs worse than the neural network-based models in identifying texts generated by different LLMs, reflecting its limitations in handling more complex language patterns.

\begin{table*}[b]
\centering
\caption{Performance Comparison of AI Models in Predicting Authorship of LLM's Texts.}
\label{tab:classification_performance}
% \vspace{-0.3cm}
\rowcolors{2}{gray!15}{white}
\resizebox{1\textwidth}{!}{%
\begin{tabular}{l|c|ccccccccccccc}
\hline
\rowcolor[HTML]{C0C0C0} 
\cellcolor[HTML]{C0C0C0} & \cellcolor[HTML]{C0C0C0} & \multicolumn{13}{c}{\cellcolor[HTML]{C0C0C0}F1-Score} \\ \cline{3-15} 
\rowcolor[HTML]{C0C0C0} 
\multirow{-2}{*}{\cellcolor[HTML]{C0C0C0}Model} & \multirow{-2}{*}{\cellcolor[HTML]{C0C0C0}Accuracy} & 
Human & GPT-3.5 & GPT-4 & Gemini-pro & Mixtral-8x7B & Mistral-7B & Meta-Llama-3-8B & Gemma-7B & Meta-Llama-3-70B & DBRX & WizardLM-2-8x22B & WizardLM-2-7B & Mixtral-8x22B \\ \hline

bert-base-cased    & \textbf{0.7095} & 0.9146 & \textbf{0.7457} & \textbf{0.7128} & \textbf{0.7970} & \textbf{0.6982} & \textbf{0.6826} & \textbf{0.6822} & \textbf{0.8762} & \textbf{0.6803} & 0.6803 & \textbf{0.7502} & \textbf{0.6393} & \textbf{0.6555} \\ \hline

DeBERTa-v3-xsmall  & 0.5790 & 0.9201 & 0.6471 & 0.5022 & 0.7204 & 0.5441 & 0.4338 & 0.6038 & 0.8062 & 0.6066 & 0.6577 & 0.4702 & 0.4651 & 0.3952 \\ \hline

DeBERTa-v3-small  & 0.6112 & 0.9199 & 0.6578 & 0.5594 & 0.7413 & 0.5825 & 0.4993 & 0.6459 & 0.8277 & 0.6122 & 0.6719 & 0.5509 & 0.4554 & 0.4495  \\ \hline

DeBERTa-v3-base    & 0.6513 & 0.9771 & 0.7012 & 0.5937 & 0.7724 & 0.6483 & 0.6130 & 0.6673 & 0.8357 & 0.5597 & \textbf{0.7231} & 0.5524 & 0.5779 &  0.5002\\ \hline

XGBoost-BoW & 0.5359 & \textbf{0.9906} & 0.5644 & 0.4696 & 0.5243 & 0.4786 & 0.4680 & 0.5218	 & 0.6168 & 0.5539 & 0.6173 & 0.5166 & 0.5297 & 0.3868  \\ \hline

\end{tabular}%
}
\captionsetup{justification=centering, singlelinecheck=false}
% \vspace{-0.4cm}
\end{table*}

\begin{tcolorbox}[colback=gray!10, colframe=black, boxsep=0.01mm, left=1mm, right=1mm, title=Summary of Key Findings for \greyball{RQ3}]
This analysis suggests that distinguishing between texts written by humans and those generated by LLMs, as well as identifying which specific LLM produced a given text, appears to be a relatively straightforward task. 
This ability to reliably identify the source of generated text not only enhances transparency but also provides valuable insights into the unique characteristics and potential biases of each LLM, contributing to more informed and ethical use of these technologies.
\end{tcolorbox}

\subsection{RQ4: Language Markers Analysis}
\begin{figure*}[h]
\begin{centering}
\subfloat[Gemini-pro.]{\includegraphics[width = 0.25\textwidth]{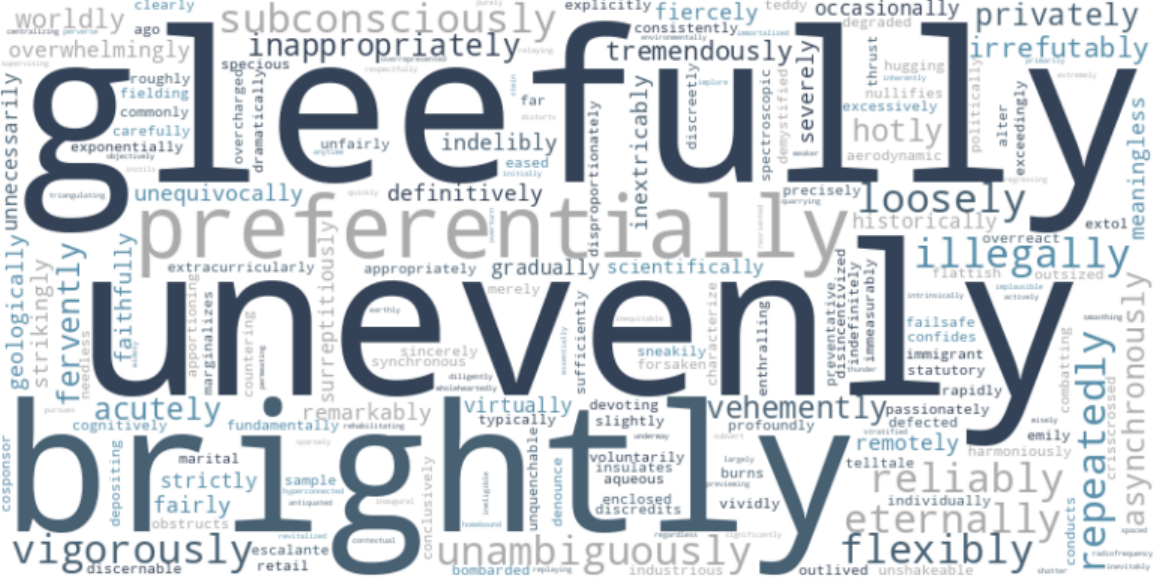}
}
\subfloat[Gemma-7B.]{\includegraphics[width = 0.25\textwidth]{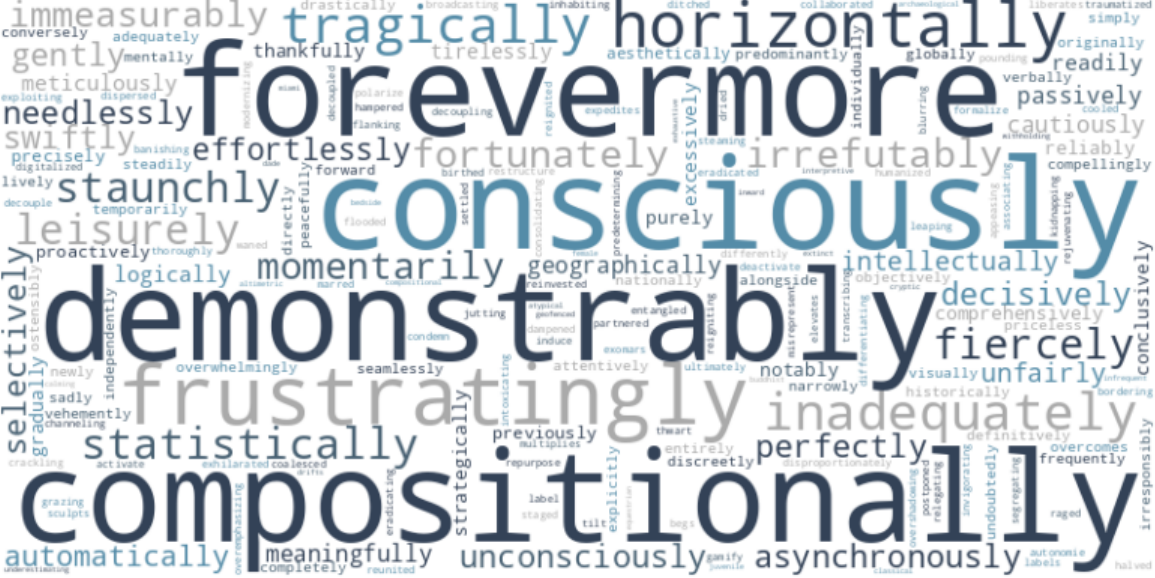}
}
\subfloat[GPT-4.]{\includegraphics[width = 0.25\textwidth]{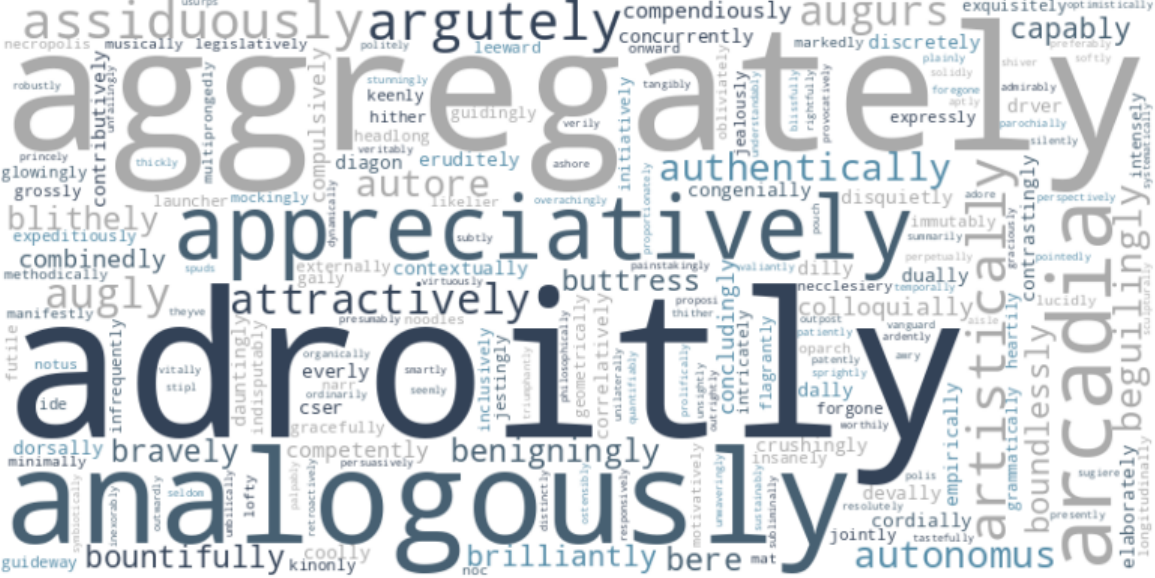}
}
\subfloat[GPT-3.5.]{\includegraphics[width = 0.25\textwidth]{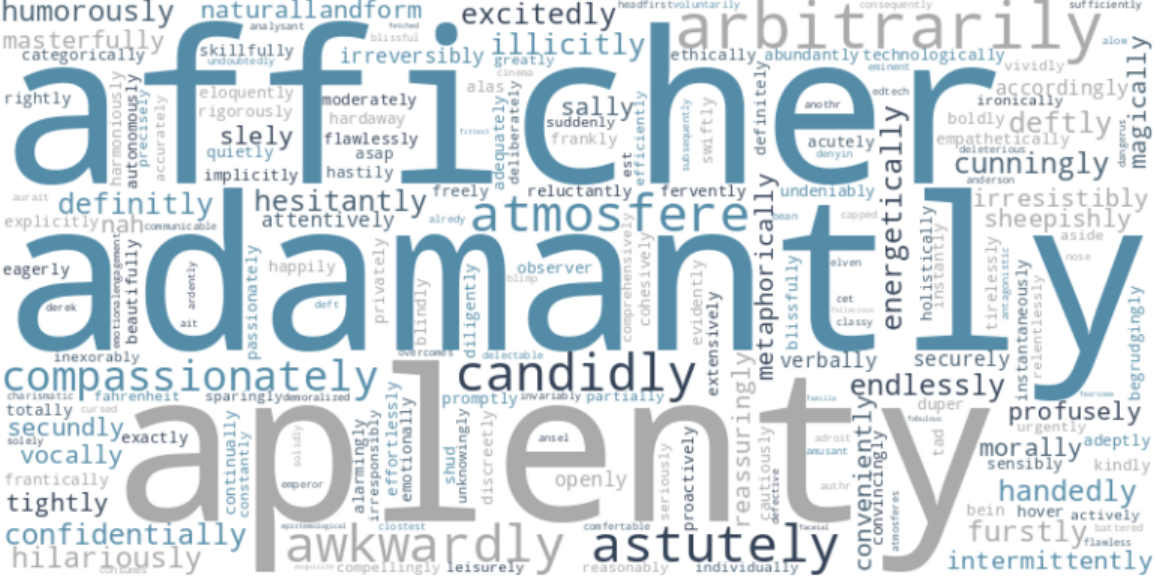}
}\\
\subfloat[Mistral-7B.]{\includegraphics[width = 0.25\textwidth]{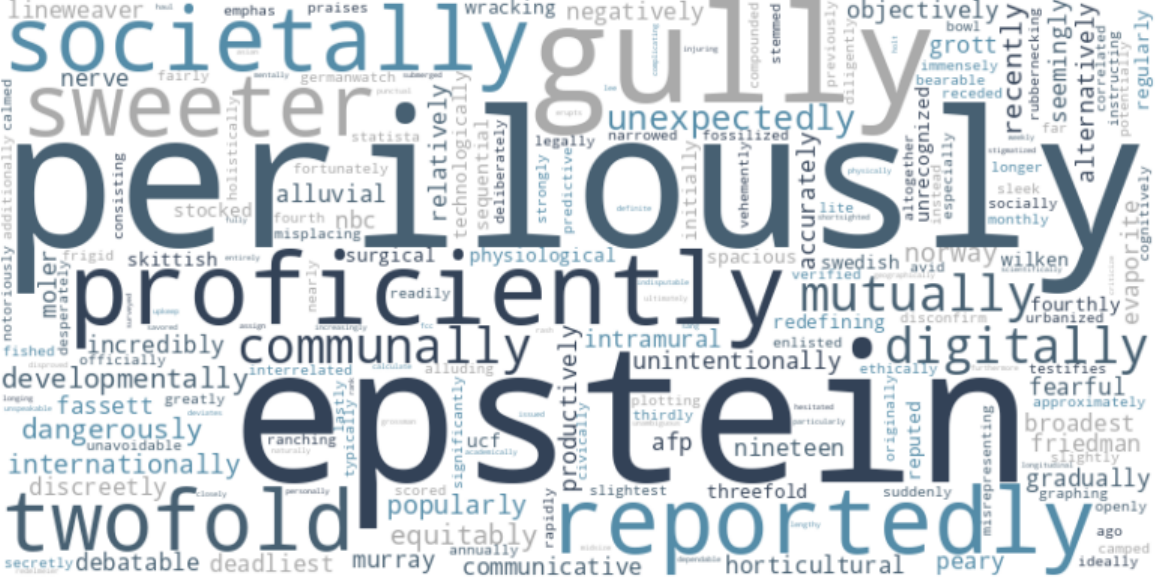}
}
\subfloat[Mixtral-8x7B.]{\includegraphics[width = 0.25\textwidth]{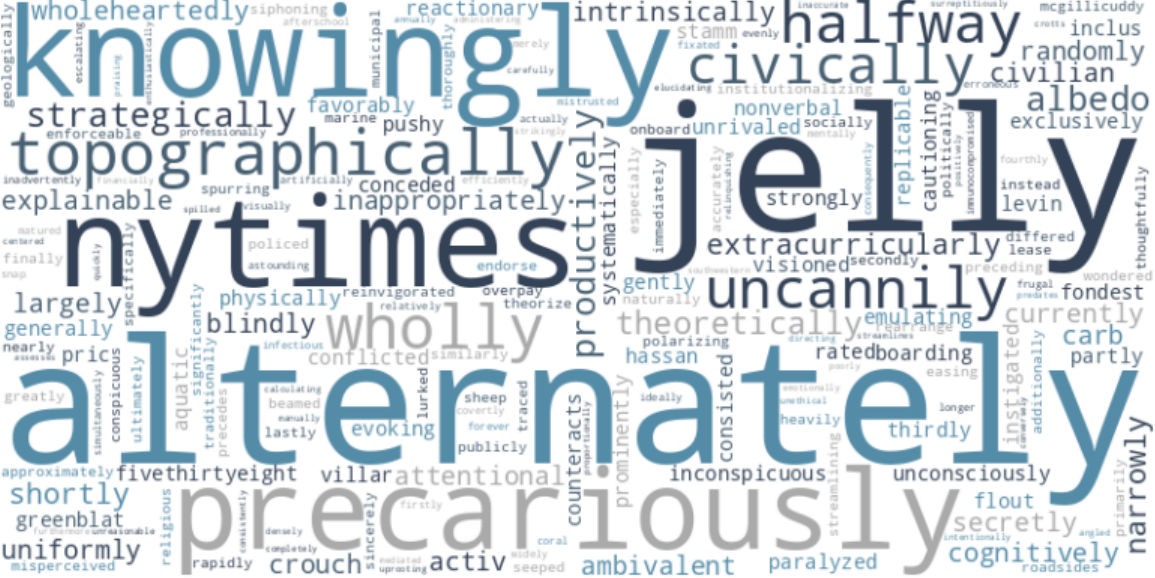}
}
\subfloat[Mixtral-8x22B.]{\includegraphics[width = 0.25\textwidth]{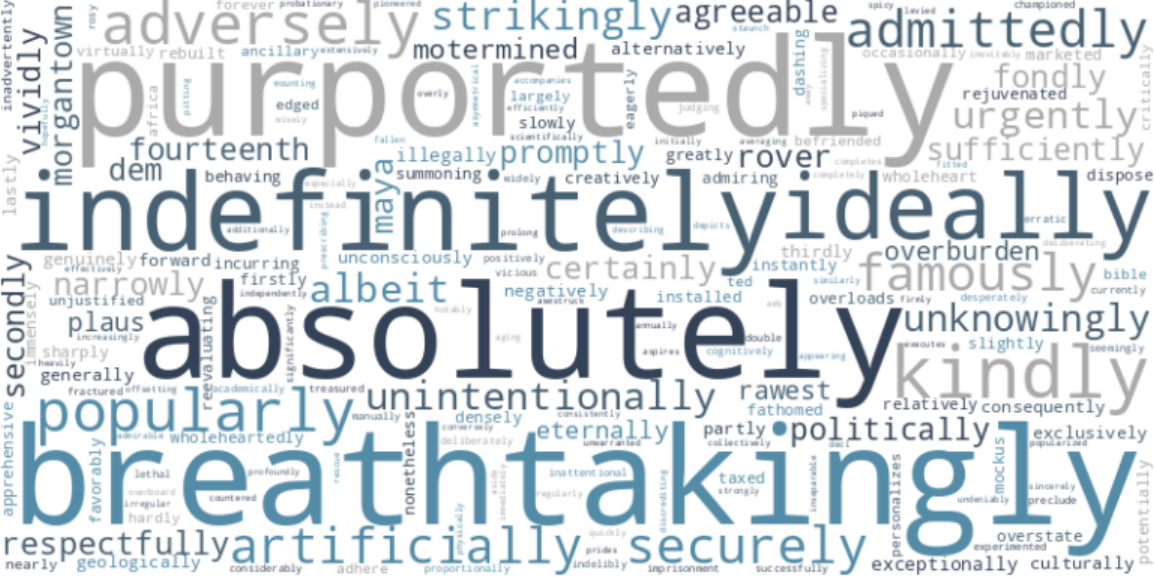}
}
\subfloat[Databricks.]{\includegraphics[width = 0.25\textwidth]{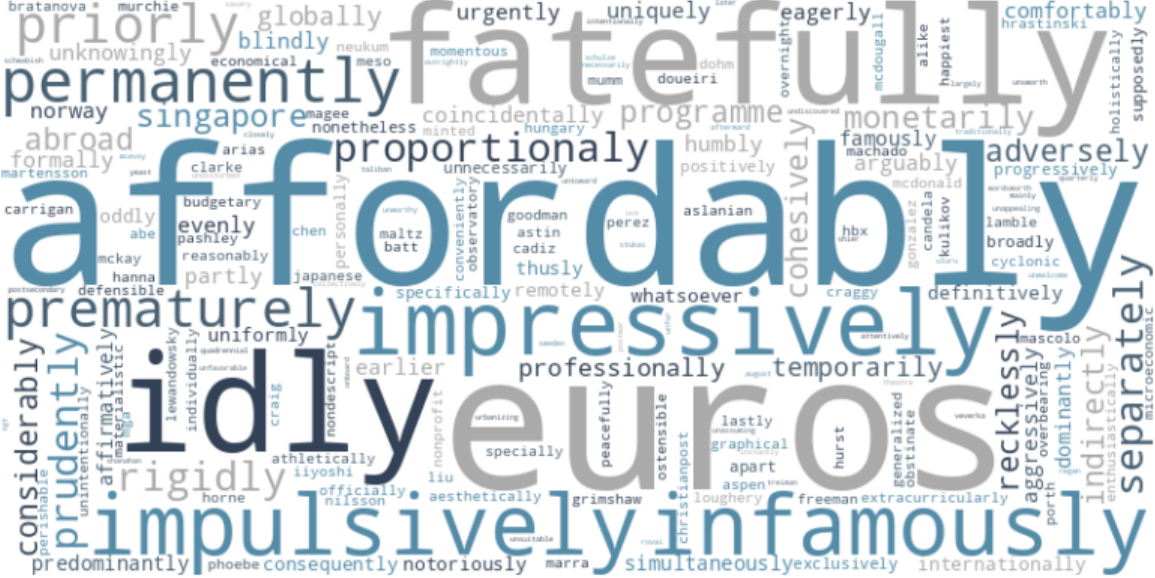}
}\\
\subfloat[Llama-3-8B.]{\includegraphics[width = 0.25\textwidth]{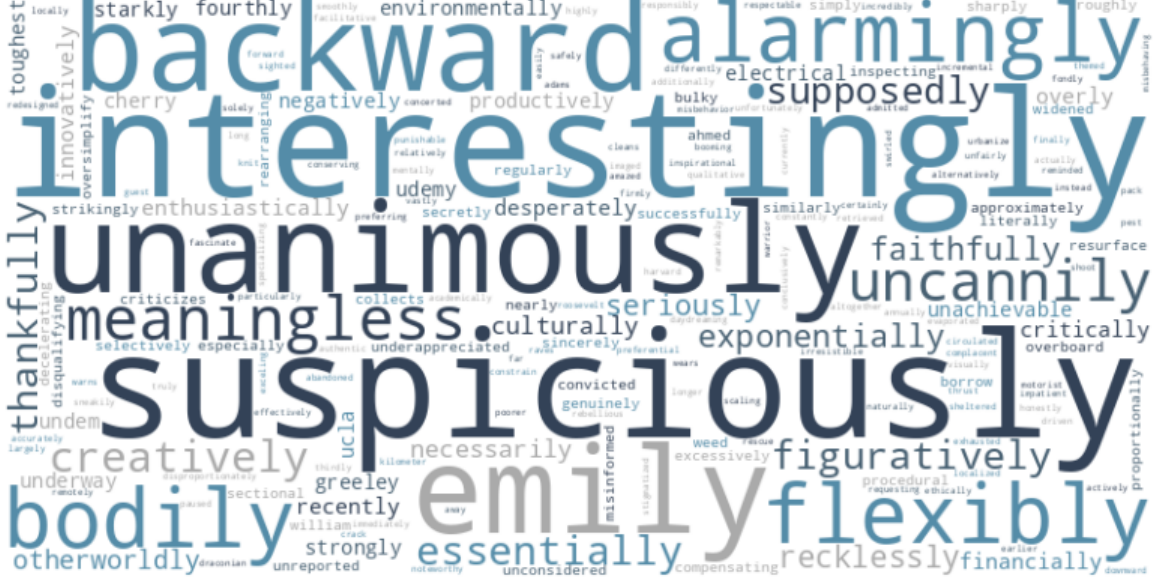}
}
\subfloat[Llama-3-70B.]{\includegraphics[width = 0.25\textwidth]{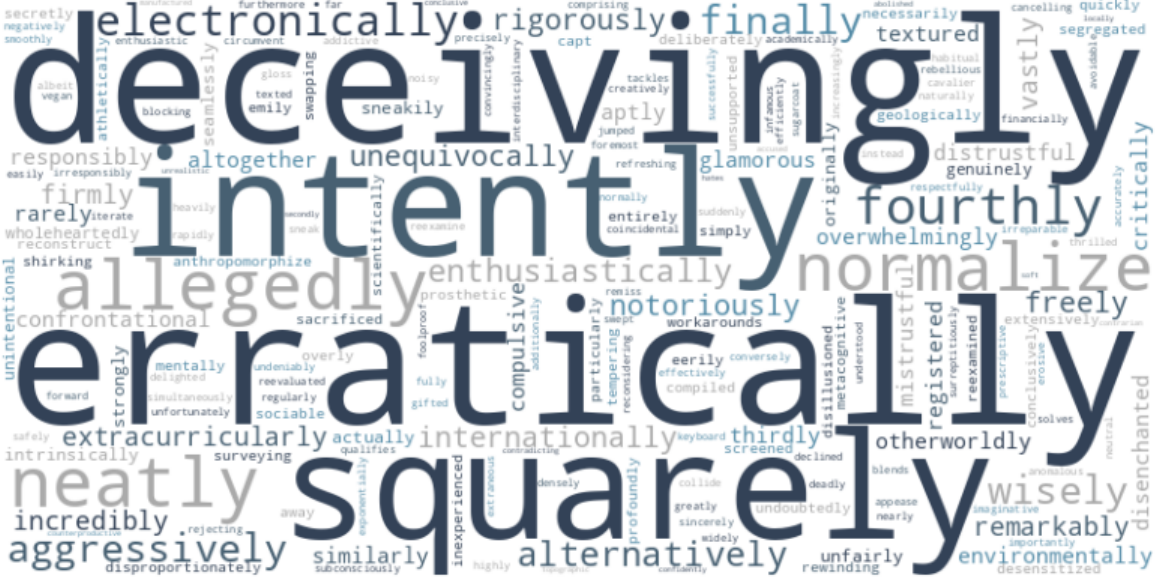}
}
\subfloat[Wizard-2-7B.]{\includegraphics[width = 0.25\textwidth]{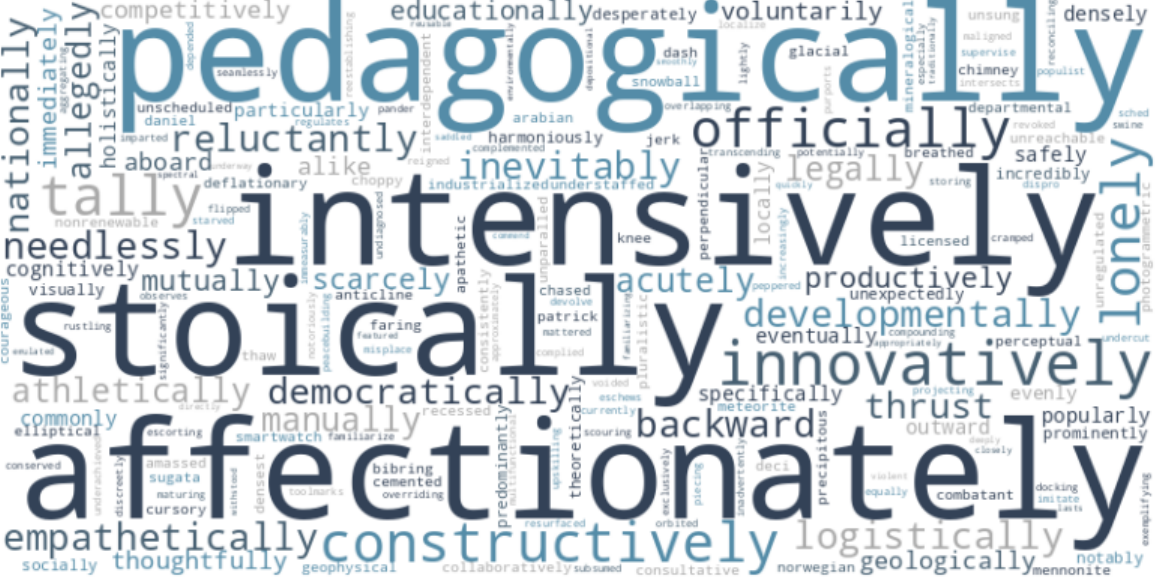}
}
\subfloat[\footnotesize{Wizard-2-8x22B}.]{\includegraphics[width = 0.25\textwidth]{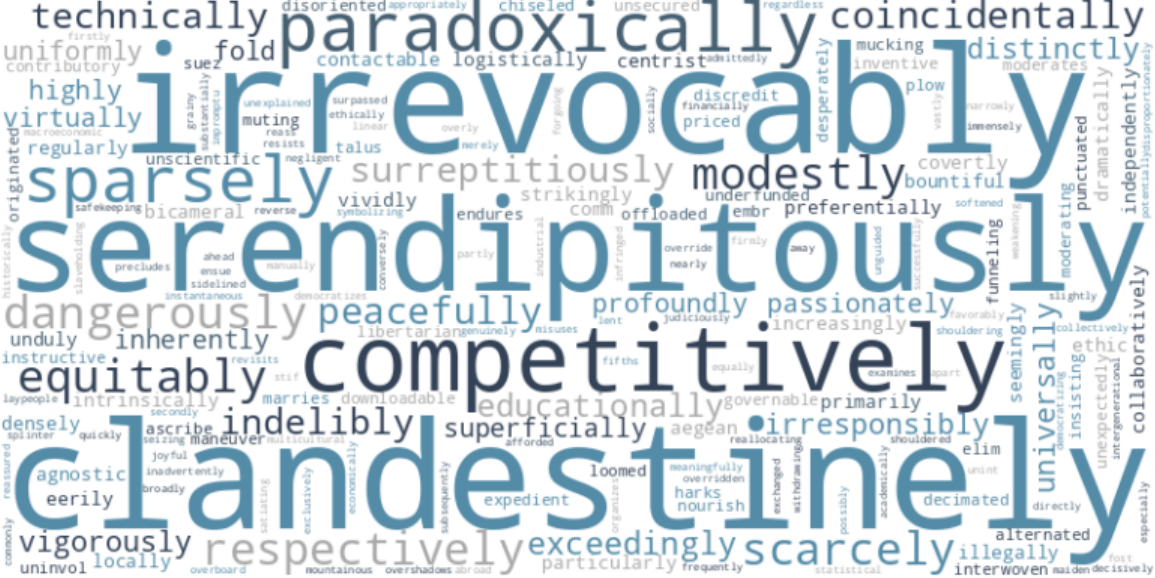}
}

\par
\end{centering}
% \vspace{-0.4cm}

\caption{Language markers.}
\label{fig:wordsclouds}

\end{figure*}

In this section, our objective is to explore the language markers, distinctive characteristics, and vocabulary exhibited by LLMs.
A general method for measuring the amount of information that a feature (i.e., a word) $x_j$ provides w.r.t. predicting a class label $y$ (i.e., the LLM generating the text or the human author) is to calculate its Point-wise Mutual Information (PMI)~\cite{church-hanks-1990-word}.
% as follows:
% \begin{equation}
%    \textit{PMI}(x_{j},y_{k})=\log\frac{P(x_{j},y_{k})}{P(x_{j})P(y_{k})}
% \end{equation}
A high PMI value indicates a more informative feature.
We leverage this information to rank and select only the most positive features (words), which are then used to generate the word clouds depicted in Figure~\ref{fig:wordsclouds}.
The language used by different language models varies significantly, even when given the same instruction. 
These observations show that each language model has distinct vocabulary and linguistic styles. This diversity highlights their unique strengths and potential applications, offering valuable insights into their capabilities.

\begin{tcolorbox}[colback=gray!10, colframe=black, boxsep=0.01mm, left=1mm, right=1mm, title=Summary of Key Findings for \greyball{RQ4}]
This analysis reveals that different LLMs exhibit distinct vocabulary and linguistic styles, as evidenced by the significant variation in language markers, making them easily recognizable and distinguishable.
\end{tcolorbox}

\subsection{RQ5: Bias and Ethics in LLMs}
In this section, we explore whether certain LLMs adhere more closely to ethical standards by effectively reducing the propagation of biased stereotypes, thereby aligning more closely with ethical guidelines in AI development and usage.

\subsubsection{Methodology}
Analyzing bias and discrimination in LLM outputs may not be a straightforward task, as these models do not explicitly exhibit racial or gender biases and stereotypes. 
Therefore, one effective approach to uncovering embedded bias in these LLMs is through an \textit{embedding-based} method. 
% This approach involves training word embeddings using data generated by each LLM, and then analyzing these embeddings to identify and quantify the embedded stereotypes and biases.

A word embedding is a representation that encodes each word $w$ as a d-dimensional vector (i.e., $w \in \mathbb{R}^d$)~\cite{NIPS2013_9aa42b31,pennington-etal-2014-glove}. 
These embeddings are trained using word co-occurrence within text corpora, leveraging paradigmatic similarity, where words with similar meanings frequently occur in similar contexts and are thus interchangeable. 
Consequently, words with related semantic meanings tend to have vectors that are close together in the embedding space. 
Moreover, the vector differences between words in these embeddings can capture the relationships between them. 
For example, in the analogy ``man is to king as woman is to $x$'', simple arithmetic on the embedding vectors reveals that the best match for $x$ is ``queen'', as the vector difference between ``man'' and ``king'' mirrors that between ``woman'' and ``queen''.
Building on the analysis by Bolukbasi et al.~\cite{Bolukbasi16}, we aim to analyze word embeddings to identify and quantify the embedded stereotypes and biases in LLM outputs.

In our experiments, we trained 50-dimensional word embeddings for each LLM's text data using the CBOW Word2Vec architecture, with a context window of 5 words and a minimum count of 1, ensuring that all words with a total frequency lower than 1 were ignored.
Once the embeddings were trained, we first evaluated them by comparing the results of various analogies against those produced by the original Word2Vec embeddings~\cite{NIPS2013_9aa42b31}. 
This allowed us to assess the quality and consistency of our trained embeddings in capturing semantic relationships between words.
We utilized the Gensim Python library to train and evaluate our word embeddings\footnote{https://radimrehurek.com/gensim/}.

\subsubsection{Identifying the Bias Subspace}
\begin{figure}[t]
\centering
\subfloat[Word pair directions to define gender and race.]{
\label{tab:Word_pair_directions}
\resizebox{0.35\textwidth}{!}{%
% Preview source code for paragraph 0

\begin{tabular}{|c|c|}
\hline 
\textbf{Gender} & \textbf{Race }  \tabularnewline
\hline 
\hline 
$\overrightarrow{\text{he}} - \overrightarrow{\text{she}}$ & $\overrightarrow{\text{white}} - \overrightarrow{\text{black}}$ \tabularnewline
$\overrightarrow{\text{him}} - \overrightarrow{\text{her}}$ & $\overrightarrow{\text{white}} - \overrightarrow{\text{asian}}$ \tabularnewline
$\overrightarrow{\text{man}} - \overrightarrow{\text{women}}$ & $\overrightarrow{\text{european}} - \overrightarrow{\text{african}}$ \tabularnewline
$\overrightarrow{\text{father}} - \overrightarrow{\text{mother}}$ & $\overrightarrow{\text{caucasian}} - \overrightarrow{\text{black}}$ \tabularnewline
$\overrightarrow{\text{boy}} - \overrightarrow{\text{girl}}$ & $\overrightarrow{\text{anglo}} - \overrightarrow{\text{afro}}$ \tabularnewline
$\overrightarrow{\text{male}} - \overrightarrow{\text{femal}}$ & $\overrightarrow{\text{white}} - \overrightarrow{\text{hispanic}}$ \tabularnewline

$\overrightarrow{\text{brother}} - \overrightarrow{\text{sister}}$ & $\overrightarrow{\text{anglo}} - \overrightarrow{\text{latino}}$ \tabularnewline
$\overrightarrow{\text{husband}} - \overrightarrow{\text{wife}}$ & $\overrightarrow{\text{white}} - \overrightarrow{\text{indigenous}}$ \tabularnewline
$\overrightarrow{\text{son}} - \overrightarrow{\text{daughter}}$ & $\overrightarrow{\text{european}} - \overrightarrow{\text{native}}$ \tabularnewline
$\overrightarrow{\text{king}} - \overrightarrow{\text{queen}}$ & $\overrightarrow{\text{white}} - \overrightarrow{\text{minority}}$ \tabularnewline
\hline 
\end{tabular}
}}
% Preview source code for paragraph 0
\hspace{0.15cm}
\subfloat[Gender-neutral words, with size indicating frequency in the data.]{\includegraphics[width = 0.45\textwidth]{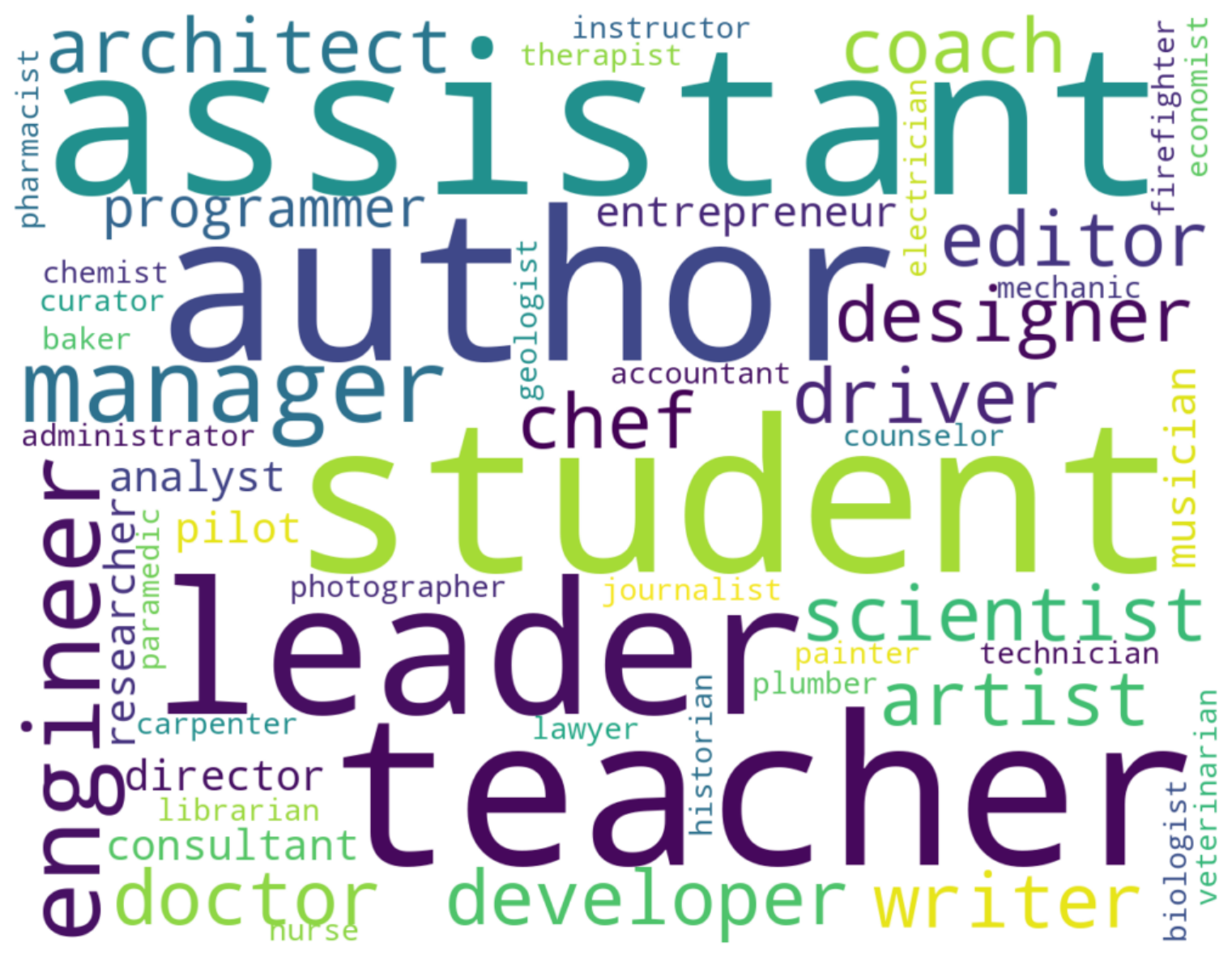}
\label{fig:Gender-neutral_words}
}
% \captionsetup{justification=centering, singlelinecheck=false}
\caption{Word pair directions to define gender and race.}
\end{figure}

In this work, we focus on two primary types of bias: \textbf{gender bias} and \textbf{racial bias}. 
As noted by Bolukbasi et al.~\cite{Bolukbasi16}, individual word pairs do not always behave as expected because a word can have multiple meanings depending on the context. 
To better estimate bias, Bolukbasi et al. proposed aggregating multiple paired comparisons to more accurately identify the bias direction subspace. 
By combining several word pair directions, such as $\overrightarrow{\text{she}} - \overrightarrow{\text{he}}$, $\overrightarrow{\text{woman}} - \overrightarrow{\text{man}}$, $\overrightarrow{\text{white}} - \overrightarrow{\text{black}}$, and $\overrightarrow{\text{european}} - \overrightarrow{\text{african}}$, we are able to identify a significant gender or racial direction $g \in \mathbb{R}^d$ that effectively captures the underlying bias in the embedding. 
Formally, the bias direction subspace $g \in \mathbb{R}^d$ is estimated as follows:

\begin{equation}
    \overrightarrow{g} = \frac{1}{|P|} \sum_{(w_1,w_2) \in P} (\overrightarrow{w_1} - \overrightarrow{w_2} )
\end{equation}

\noindent where $P$ is the list of word pair directions shown in Figure~\ref{tab:Word_pair_directions}, and $(w_1, w_2) $ is a pair of words in $P$.
The direction represented by $g$ allows us to quantify direct biases in word associations, offering a more comprehensive understanding of how bias manifests in the embeddings.

\subsubsection{Estimating Direct Bias }

To measure direct bias, we first defined a list $N$ of 50 words that are expected to be gender-neutral, as illustrated in Figure~\ref{fig:Gender-neutral_words}. 
Then, given a gender-neutral word from $N$, and the gender direction $g$ learned earlier, we estimate the direct bias of an embedding using cosine similarity, as suggested in~\cite{Bolukbasi16}:

\begin{equation}
    b_w = cos(\overrightarrow{w},\overrightarrow{g})
    \label{eq:bias}
\end{equation}

\noindent where a positive value of $b_w$ indicates that $w$ is more strongly associated with male, white, European, or Caucasian, while a negative value of $b_w$ suggests a stronger association with female, Black, African, or Asian.
Finally, we estimate the overall direct bias of the embeddings as follows:
\begin{equation}
    DirectBias = \frac{1}{|N|} \sum_{w \in N} |b_w|
    \label{eq:direcbias}
\end{equation}

\noindent where a lower value of $DirectBias$ indicates a lower level of bias.

\begin{figure}[t]
\begin{centering}
\includegraphics[width = 0.55\textwidth]{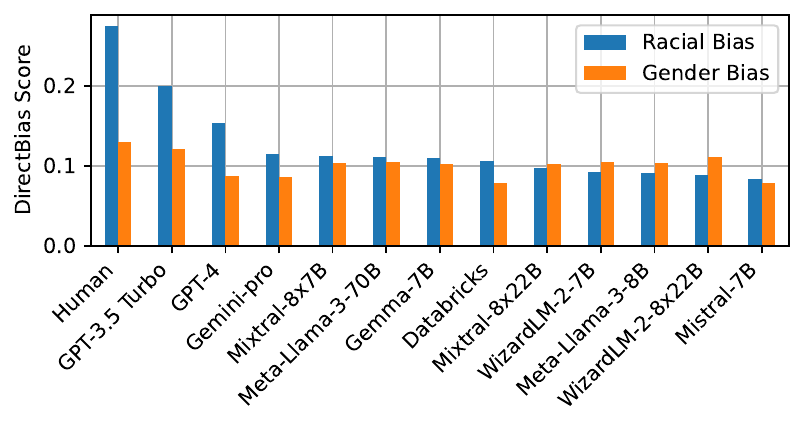}
\par
\end{centering}
% \vspace{-0.43cm}
\caption{DirectBias scores.}
\label{fig:direct_bias}
\end{figure}

\begin{figure}[t]
\begin{centering}
\includegraphics[width = 0.55\textwidth]{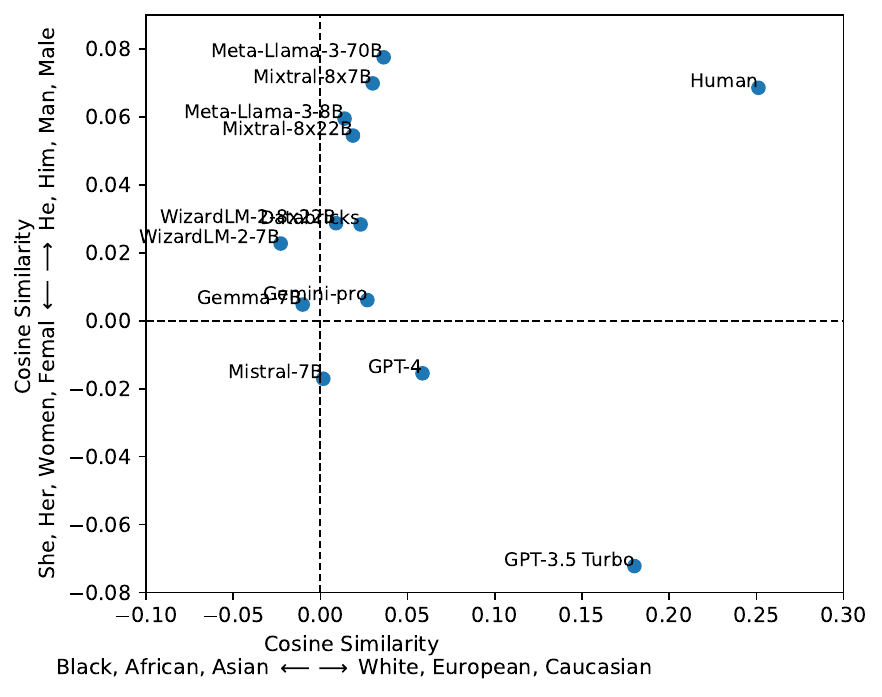}
\par
\end{centering}
% \vspace{-0.43cm}
\caption{Two-dimensional scatter plot of the association score between each LLM and each bias dimension.}
\label{fig:llms_bias}
% \vspace{-0.43cm}
\end{figure}

\begin{figure*}[t]
\begin{centering}
\subfloat[Gemini-pro.]{\includegraphics[width = 0.25\textwidth]{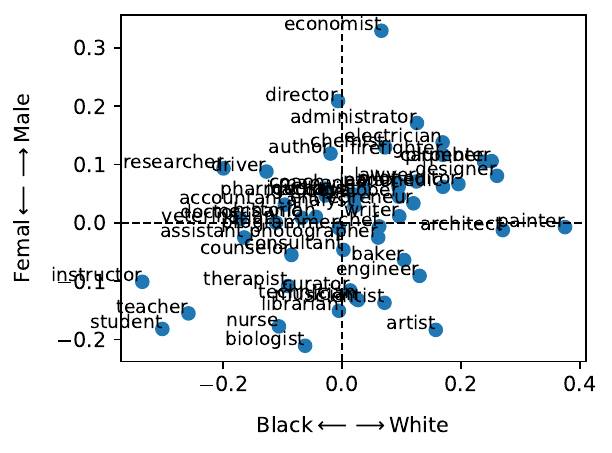}
}
\subfloat[Gemma-7B.]{\includegraphics[width = 0.25\textwidth]{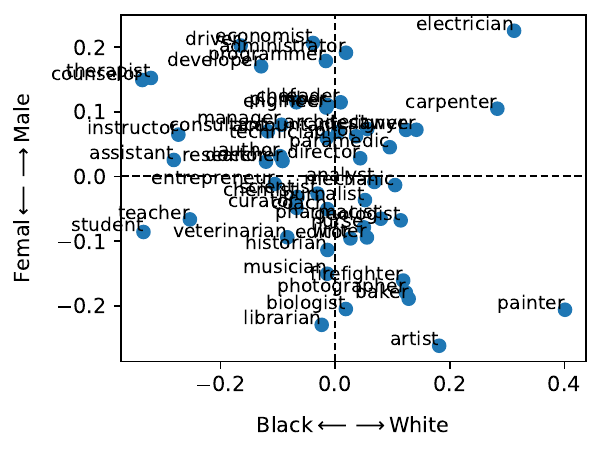}
}
\subfloat[GPT-4.]{\includegraphics[width = 0.25\textwidth]{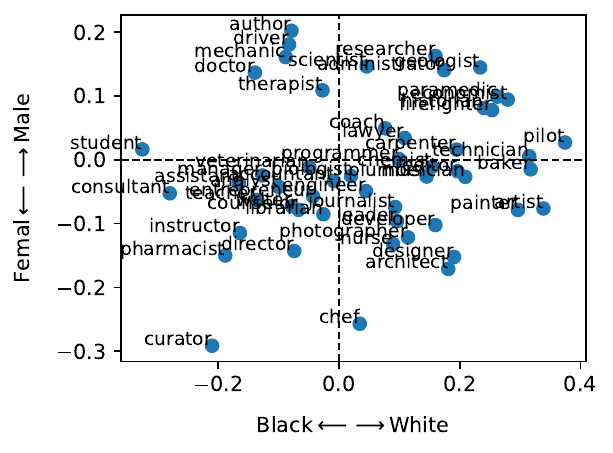}
}
\subfloat[GPT-3.5.]{\includegraphics[width = 0.25\textwidth]{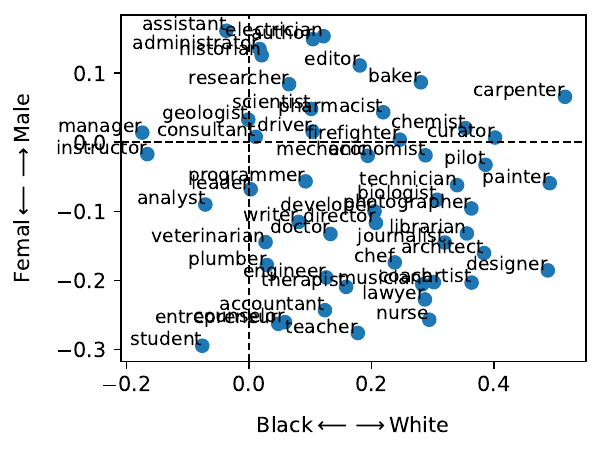}
}
\par
\subfloat[Mistral-7B.]{\includegraphics[width = 0.25\textwidth]{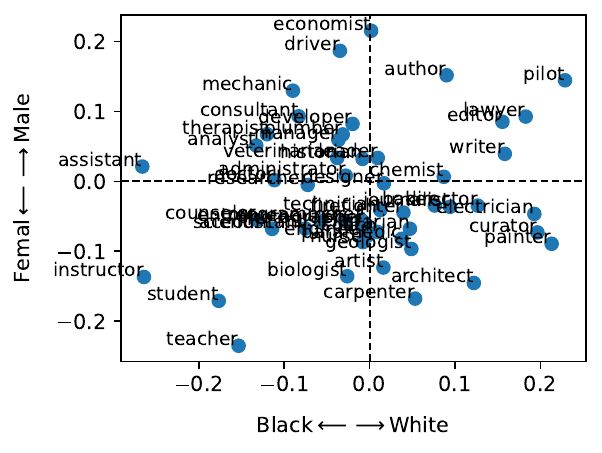}}
\subfloat[Mixtral-8x7B.]{\includegraphics[width = 0.25\textwidth]{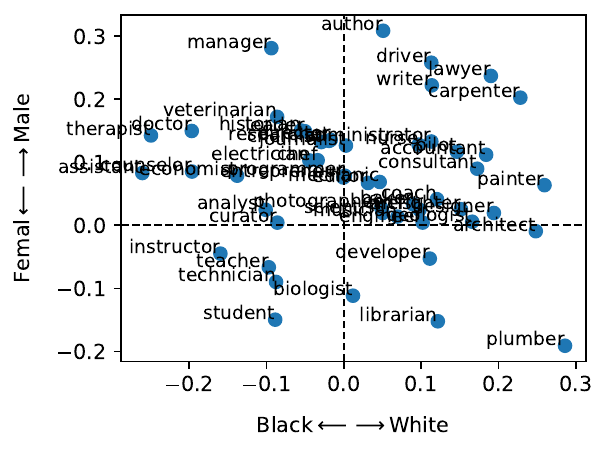}
}
\subfloat[Mixtral-8x22B.]{\includegraphics[width = 0.25\textwidth]{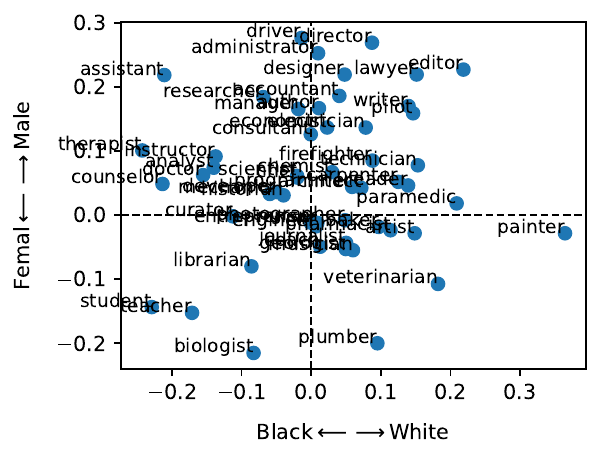}
}
\subfloat[Databricks.]{\includegraphics[width = 0.25\textwidth]{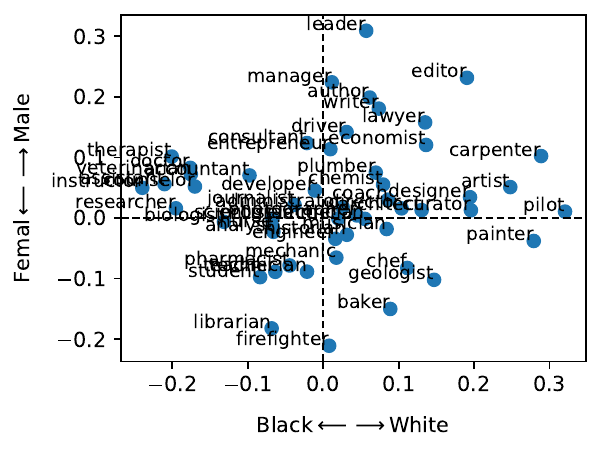}
}
\par
\subfloat[Llama-3-8B.]{\includegraphics[width = 0.25\textwidth]{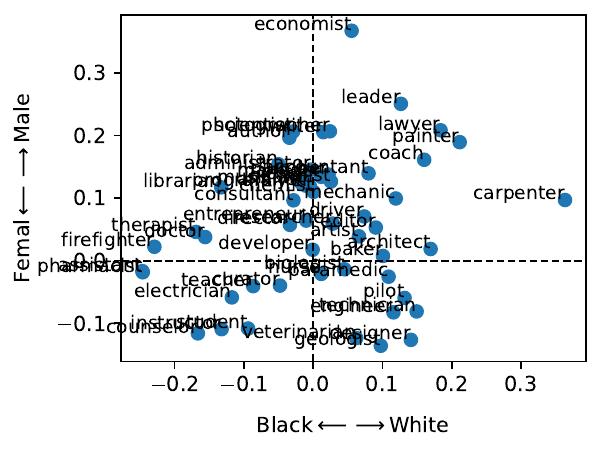}}
\subfloat[Llama-3-70B.]{\includegraphics[width = 0.25\textwidth]{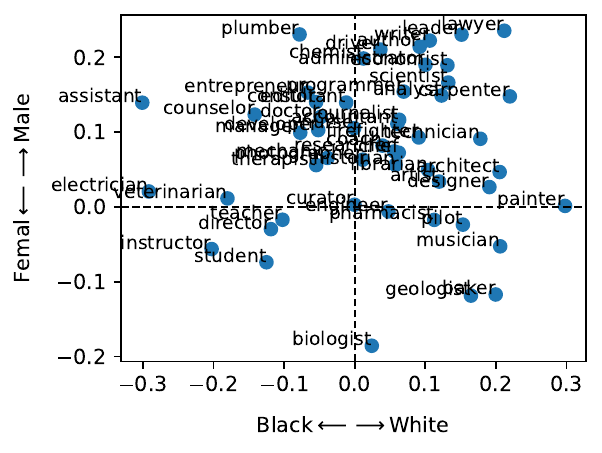}
}
\subfloat[Wizard-2-7B.]{\includegraphics[width = 0.25\textwidth]{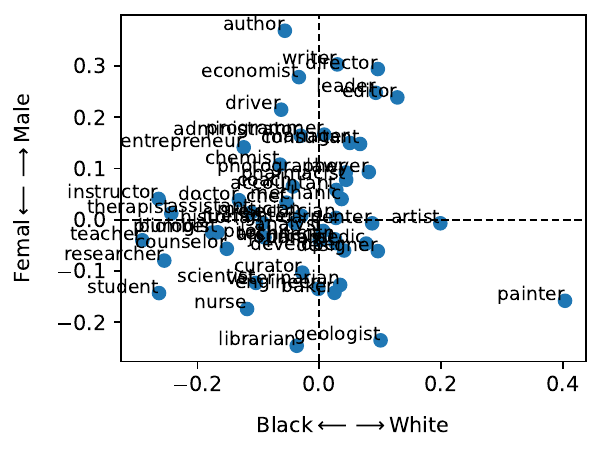}
}
\subfloat[Wizard-2-8x22B.]{\includegraphics[width = 0.25\textwidth]{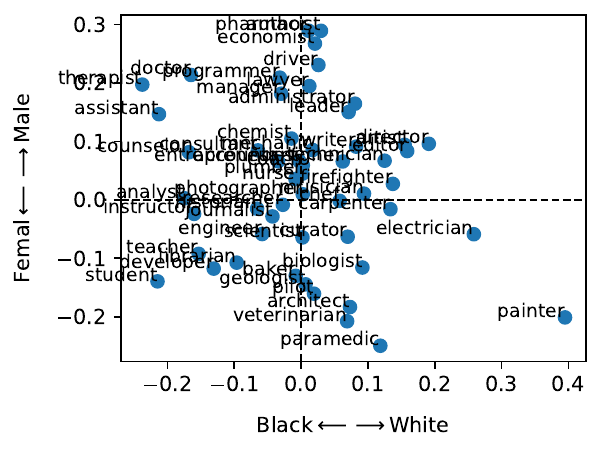}
}
\end{centering}
% \vspace{-0.3cm}
\caption{Two-dimensional scatter plot of the association score between each occupation of Figure~\ref{fig:Gender-neutral_words} and each bias dimension.}
\label{fig:bias_models_occupations}
% \vspace{-0.4cm}
\end{figure*}

\subsubsection{Bias assessment}
Figure~\ref{fig:direct_bias} shows the DirectBias score, calculated using Equation~\ref{eq:direcbias}, for each model across different bias dimensions. At first glance, it is apparent that, overall, all models exhibit a relatively similar level of latent gender bias. However, human-generated texts, along with GPT-3.5 and GPT-4, appear to contain the most latent racial bias.

To further analyze which models are more strongly associated with specific bias dimensions, we refer to Figure~\ref{fig:llms_bias}, which presents the average bias scores calculated using Equation~\ref{eq:bias} for each model across various bias dimensions. Here, we observe that some models show stronger associations with particular biases. For instance, human-generated texts tend to have a strong association with white males, whereas models like GPT-3.5 and GPT-4 exhibit a stronger latent association with white females, suggesting that these models lean more towards feminist viewpoints compared to others. Additionally, we note that Gemma-7B and Gemini-pro are positioned closer to the origin (0,0), indicating that they are the most balanced models in terms of bias.

Finally, Figure~\ref{fig:bias_models_occupations} presents a two-dimensional scatter plot illustrating the association score between each occupation from Figure~\ref{fig:Gender-neutral_words} and ou two bias dimensions. 
Several interesting stereotypes emerge from the data. 
For example, in Databricks, a leader and a manager are more strongly associated with being white males, whereas in Gemini-pro, a nurse is more closely associated with being a Black female.

\begin{tcolorbox}[colback=gray!10, colframe=black, boxsep=0.01mm, left=1mm, right=1mm, title=Summary of Key Findings for \greyball{RQ5}]
We note that all models exhibit relatively similar levels of latent gender bias in general. 
However, certain models demonstrate stronger associations with specific bias dimensions. 
For instance, GPT-3.5 and GPT-4 show a stronger association with females, suggesting a tendency towards feminist viewpoints. 
Also, in terms of racial bias, GPT-3.5 and GPT-4 display the highest levels of latent racial bias, particularly associating leadership roles with white males. 
Finally, models like Gemma-7B and Gemini-pro are identified as the most balanced models.
\end{tcolorbox}

% \newpage
\section{Conclusion and Future work}
In conclusion, our analysis sheds light on critical aspects of Large Language Models (LLMs). The observed low similarity within LLMs, distinctive inter-LLM writing styles, varying degrees of variance in text generation, successful classification outcomes, and discernible language markers underscore the nuanced and complex nature of LLM behavior. 
Moreover, we demonstrate that LLMs differ in their associations with gender and racial biases, with models like GPT-3.5 and GPT-4 showing a stronger tendency towards feminist viewpoints and latent racial bias. Balanced models such as Gemma-7B and Gemini-pro are identified as exhibiting the least bias overall. 
These findings contribute to a deeper understanding of LLM capabilities, providing valuable insights for future advancements in natural language processing and model interpretability.

Future work involves exploring explainability techniques, where the focus extends beyond  detecting whether a text is authored by a human or generated by an LLM. 
The aim is to explore and articulate the reasons behind the model's predictions, providing a more comprehensive understanding of the decision-making process. 

\subsubfour{Acknowledgement:}
all data used in this paper will be made publicly available and released following the publication of this paper, in accordance with the applicable data sharing policies and guidelines in place.

% \clearpage

\bibliographystyle{unsrt}
% \balance
\bibliography{biblio.bib}

\end{document}